\documentclass{article}

\PassOptionsToPackage{numbers, compress}{natbib}

\usepackage[main,final]{neurips_2026}

\usepackage[utf8]{inputenc} %
\usepackage[T1]{fontenc}    %
\usepackage{hyperref}       %
\usepackage{url}            %
\usepackage{booktabs}       %
\usepackage{amsfonts}       %
\usepackage{nicefrac}       %
\usepackage{microtype}      %
\usepackage{xcolor}         %

\usepackage{graphicx}
\usepackage{subcaption}
\usepackage{adjustbox}
\usepackage{multirow}
\usepackage{makecell}      %
\usepackage{threeparttable}
\usepackage{amsmath}
\usepackage{cleveref}
\crefname{figure}{Fig.}{Figs.}
\Crefname{figure}{Fig.}{Figs.}
\crefname{table}{Tab.}{Tabs.}
\Crefname{table}{Tab.}{Tabs.}
\crefname{section}{Sec.}{Secs.}
\Crefname{section}{Sec.}{Secs.}
\crefname{subsection}{Sec.}{Secs.}
\Crefname{subsection}{Sec.}{Secs.}
\crefname{subsubsection}{Sec.}{Secs.}
\Crefname{subsubsection}{Sec.}{Secs.}
\crefname{equation}{Eq.}{Eqs.}
\Crefname{equation}{Eq.}{Eqs.}
\usepackage{arydshln}
\usepackage{wrapfig}

\newcommand{\fullmark}{\textcolor{green!55!black}{\Large\checkmark}} %
\newcommand{\nonemark}{\textcolor{red!70!black}{\Large\(\times\)}}   %
\newcommand{\halfmark}{\rule[0.6ex]{1em}{1.1pt}}
\newcommand\blfootnote[1]{%
\begingroup
\renewcommand\thefootnote{}\footnote{#1}%
\addtocounter{footnote}{-1}%
\endgroup
}

\title{Tri-Prompting: Controllable Video Generation with Scene, Subject, and Motion Prompts}

\author{%
  \textbf{Zhenghong~Zhou}$^{1,2*}$ \enspace
  \textbf{Xiaohang~Zhan}$^{1}$ \enspace
  \textbf{Zhiqin~Chen}$^{1}$ \enspace
  \textbf{Soo~Ye~Kim}$^{1}$ \enspace
  \textbf{Nanxuan~Zhao}$^{1}$ \\
  \textbf{Haitian~Zheng}$^{1}$ \enspace
  \textbf{Qing~Liu}$^{1}$ \enspace
  \textbf{He~Zhang}$^{1}$ \enspace
  \textbf{Zhe~Lin}$^{1}$ \enspace
  \textbf{Yuqian~Zhou}$^{1\ddagger}$ \enspace
  \textbf{Jiebo~Luo}$^{2\ddagger}$ \\[2pt]
  $^{1}$Adobe Research \qquad $^{2}$University of Rochester \qquad $^{\ddagger}$Equal advising \\[2pt]
  \href{https://zhouzhenghong-gt.github.io/Tri-Prompting-Page/}{Project Page}
}

\hypersetup{
  pdftitle={Tri-Prompting: Controllable Video Generation with Scene, Subject, and Motion Prompts},
  pdfauthor={Zhenghong Zhou, Xiaohang Zhan, Zhiqin Chen, Soo Ye Kim, Nanxuan Zhao, Haitian Zheng, Qing Liu, He Zhang, Zhe Lin, Yuqian Zhou, Jiebo Luo}
}

\begin{document}

\maketitle
\blfootnote{$^{*}$Work done during an internship at Adobe.}
\begin{center}
    \vspace{-5mm}
    \centering
    \includegraphics[width=0.85\linewidth]{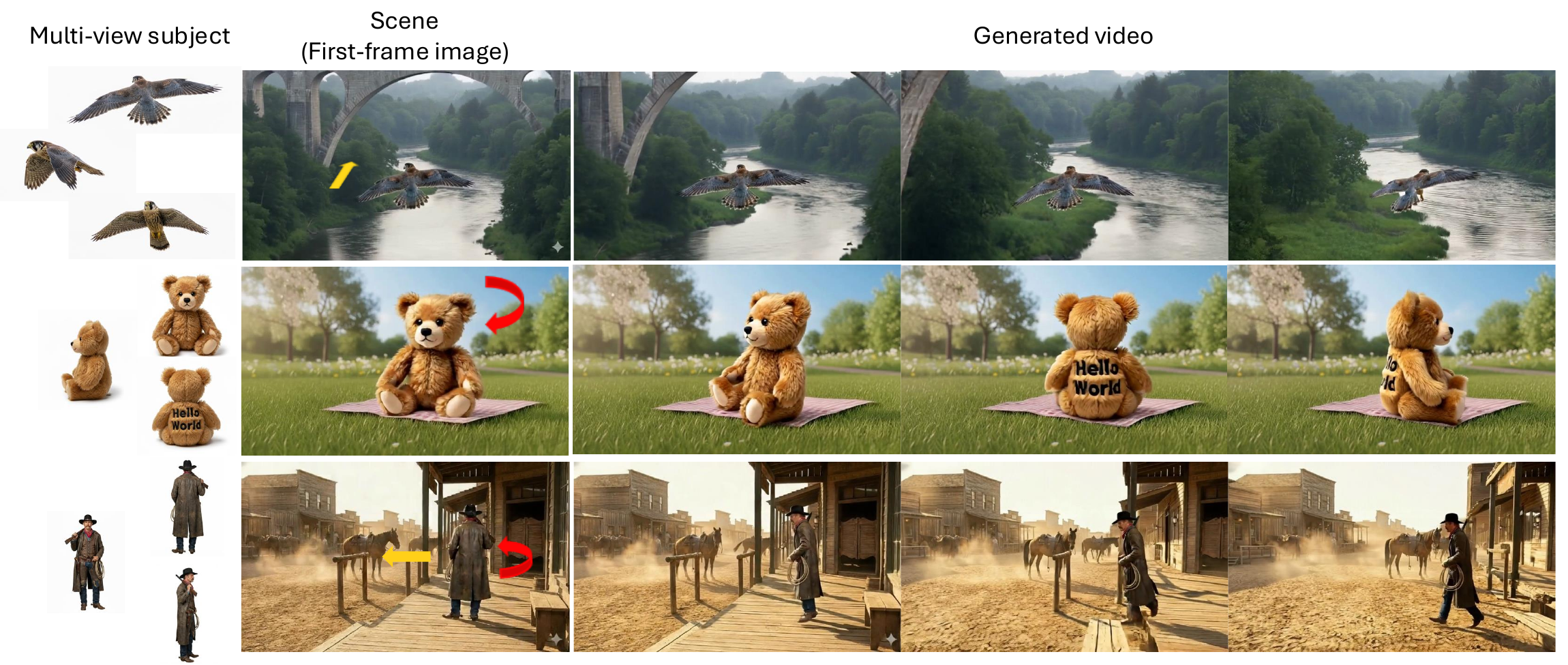}
    \vspace{-2mm}
    \captionof{figure}{Tri-Prompting integrates scene control, multi-view subject control, and scene–subject motion control within one video diffusion model. Users can select a 3D-aware subject, insert or manipulate it in diverse scenes, and specify camera and subject motion through an offline control interface with keyboard, while maintaining appearance consistency with reference images. The figure shows scene/camera-only, subject-only, and joint scene-subject motion control. \textcolor{yellow!80!black}{Yellow arrow →} indicates camera trajectory, and \textcolor{red}{Red arrow →} indicates subject motion.}
    \label{fig:teaser}
\end{center}

\vspace{-1.5mm}
\begin{abstract}
\vspace{-1.5mm}

Recent video diffusion models achieve strong visual quality and temporal coherence, but still lack coordinated control over scene layout, customized subject identity, and camera/subject motion.
We study controllable video generation from three prompts: a first-frame scene image, 3D-aware multi-view subject references, and a motion-driving signal.
This setting is challenging because background regions are often trackable under camera motion, while foreground subjects can rotate, self-occlude, and reveal new regions that require identity-consistent appearance.
We introduce Tri-Prompting, a video diffusion framework that combines multi-view subject conditioning with dual-conditioned motion control.
Tri-Prompting uses XYZ tracking points for visible background motion and low-resolution RGB proxies for foreground subject pose, allowing the model to recover fine appearance from multi-view references while retaining flexibility for plausible subject-scene interactions.
An inference-time ControlNet scale schedule further balances motion controllability and visual realism.
We evaluate Tri-Prompting against specialized baselines: DaS for motion reconstruction and Phantom for subject-driven generation.
Tri-Prompting achieves competitive or improved reconstruction quality, stronger multi-view identity preservation, and better 3D consistency, while enabling 3D-aware subject insertion and in-scene manipulation.

\end{abstract}

\section{Introduction}
\label{sec:intro}
\vspace{-1mm}

Recent video diffusion models deliver remarkable visual quality and temporal coherence~\cite{wan,cogvideox}.
However, video creation is naturally organized around three questions: where the story happens, who appears in it, and how the camera and subject move.
In this work, we instantiate these three factors as scene, subject, and motion prompts: a first-frame scene image, multi-view subject references, and a motion-driving signal.
The goal is to generate a video that preserves the scene layout and 3D-aware subject identity while following the specified camera and subject motion.
For example, starting from a single scene image, a creator may insert a new 3D-aware subject or manipulate an existing one, move the camera, and turn the subject around.

\begin{table}[htbp]
    \centering
    \small
    \setlength{\tabcolsep}{6pt}
    \caption{Factorized controllability.}
    \label{tab:comparison}
    \begin{adjustbox}{max width=\linewidth}
      \begin{tabular}{@{}lccc@{}}
        \toprule
        & \makecell{\textbf{Where}\\Scene (first frame)}
        & \makecell{\textbf{Who}\\Multi-views of subject}
        & \makecell{\textbf{How}\\Cam. / Subj. motion} \\
        \midrule
        I2V (Wan)~\cite{wan}             & \fullmark & \nonemark & \nonemark \\
        Phantom~\cite{phantom}               & \nonemark & \halfmark & \nonemark \\
        DaS~\cite{das}                   & \fullmark & \halfmark & \halfmark \\
        \midrule
        \textbf{Ours (Tri-Prompting)} & \fullmark & \fullmark & \fullmark \\
        \bottomrule
      \end{tabular}
    \end{adjustbox}
    \vspace{2pt}
    \parbox{\linewidth}{\footnotesize
    \textit{Notes.} \fullmark{} denotes full support, \nonemark{} denotes no support, and \halfmark{} denotes partial or limited support.
    Specialized methods only handle isolated factors.
    \emph{Phantom} generates videos from single-view subjects (\halfmark); \emph{DaS} enables motion control video generation (\halfmark) but is limited to the first frame, and fails to maintain multi-view appearance consistency.
    In contrast, \textbf{Tri-Prompting} covers all three dimensions.}
  \end{table}

Existing methods cover only part of this tri-prompt setting.
Motion-guided methods such as MotionPrompting~\cite{motionprompting} and DaS~\cite{das} provide trajectory control, but rely mainly on first-frame appearance and struggle to preserve customized identity under large view changes.
Subject-driven methods such as Phantom~\cite{phantom} preserve subject appearance from references, but lack camera and subject motion control.
Thus, existing methods do not directly support first-frame scene control, multi-view subject identity, and motion control in one generation framework.

This setting poses two challenges.
First, background and foreground motion have different trackability.
Background regions are often trackable under camera motion, making tracking points effective for scene control.
In contrast, foreground subjects can rotate, self-occlude, or deform, causing regions to disappear or become newly visible, which makes tracking-point controls unreliable for subject motion.
Second, the subject should be represented as a complete object that can be viewed from different angles.
A single reference image cannot reliably specify the side or back appearance, making multi-view references important for identity preservation under large viewpoint changes.

We introduce Tri-Prompting, a video diffusion framework for jointly controlling scene, subject, and motion.
To address the different trackability of background and foreground motion, we use separate motion representations: XYZ trajectories for visible background surfaces and downsampled RGB grids for foreground subjects whose tracking points may disappear under rotation or self-occlusion.
The low-resolution RGB proxy encodes 3D-aware subject pose without over-specifying fine appearance details, allowing the model to recover side and back appearance from multi-view references while retaining flexibility to synthesize realistic subject-scene interactions.
We implement this design in two stages: the first stage learns to fuse the first-frame scene and multi-view subject references, and the second stage trains a ControlNet module for dual-conditioning motion control.
At inference time, we further use a ControlNet scale schedule to reduce over-constraining from coarse motion cues and balance motion controllability with visual realism.

We evaluate Tri-Prompting through comparisons with the closest specialized baselines.
For motion controllability, we compare with DaS on video reconstruction and show that dual conditioning improves reconstruction quality, especially in perceptual similarity.
For multi-view subject-to-video generation, we compare with Phantom and observe consistent gains in video quality, multi-view identity similarity, and 3D consistency.
Beyond these evaluations, we demonstrate 3D-aware subject insertion and in-scene manipulation workflows, where users can specify scene, subject, and motion prompts within the same framework.
We fine-tune Tri-Prompting with 11k tuples ($<$7 hours of video) for fewer than 5k steps.

In summary, our contributions are as follows:
\begin{itemize}
\item \textbf{Tri-prompting for controllable video generation.}
We propose Tri-Prompting, a single video diffusion framework that integrates scene, subject, and motion prompts for controllable video generation.

\item \textbf{Dual-conditioning motion control with multi-view subject consistency.}
We design a motion control module that combines XYZ trajectories for background motion and low-resolution RGB proxies for foreground subject motion, enabling decoupled scene/subject control while preserving identity with multi-view references.

\item \textbf{3D-aware subject insertion / manipulation workflows and competitive results.}
Tri-Prompting enables subject insertion and in-scene manipulation workflows, and achieves competitive or improved performance against the closest specialized baselines, including DaS for motion reconstruction and Phantom for subject identity preservation.
\end{itemize}

\vspace{-2mm}
\section{Related work}\label{sec:relatedwork}
\vspace{-2mm}
\noindent\textbf{Text-to-Video and Image-to-Video Diffusion Models.}
Text-to-video (T2V) and image-to-video (I2V) diffusion models have progressed rapidly.
Early video diffusion models~\cite{he2022latent, wu2023tune, guo2023animatediff, blattmann2023stable} extended text-to-image latent diffusion models by adding temporal layers to UNet backbones.
Recent transformer-based video diffusion models~\cite{cogvideox, brooks2024video, kong2024hunyuanvideo, ma2025step, wan, polyak2024movie} represent videos as spatio-temporal tokens and use full attention to improve visual quality, temporal coherence, and motion diversity.
Despite these advances, text or image prompts alone remain insufficient for precise control over scene layout, subject identity, and motion.

\noindent\textbf{Motion-Guided Video Generation.}
Motion-guided video generation uses spatial or temporal cues such as pose, boxes, depth, camera trajectories, and tracking points~\cite{ma2024follow, xing2024make, xing2024tooncrafter, wang2024drivedreamer, chen2025echomimic, shi2024motion, yu2024viewcrafter, das, wang2025epic, motionprompting, latentreframe}.
Trajectory- and tracking-based methods are closely related to our motion prompt: MotionPrompting~\cite{motionprompting} uses 2D motion trajectories, while DaS~\cite{das} uses 3D tracking points to control visible scene motion.
These methods are effective for moving visible content, but they do not address 3D-aware subject control: when a customized subject rotates or self-occludes, existing tracking points disappear, while newly visible surfaces have no corresponding tracks.
Direct-a-Video~\cite{yang2024direct} and MotionBooth~\cite{wu2024motionbooth} further support decoupled object and camera control, but mainly through 2D box controls rather than 3D-aware subject motion that reveals new views of a customized object.
They also do not use a first-frame scene image for scene preservation.
Frame In-N-Out~\cite{wang2026frame} controls object entry and exit on an expanded canvas using 2D trajectories.
CustomX~\cite{wang2025customx} animates customized characters in a 3D Gaussian scene with text-guided actions and camera control.
In contrast, Tri-Prompting uses a first-frame scene image, multi-view subject references, and separate background/foreground motion anchors to enable 3D-aware motion such as subject rotation.

\noindent\textbf{Subject-Driven Video Customization.}
Subject-driven video customization aims to preserve the appearance of user-provided subjects across generated videos~\cite{jiang2024videobooth, zhuang2024vlogger, fei2025skyreels, phantom, cai2025omnivcus, ju2025editverse}.
Single-view references provide only partial appearance, making side and back views ambiguous under large viewpoint changes.
Recent methods, including 3DreamBooth~\cite{ko20263dreambooth} and MV-S2V~\cite{song2026mv}, use multi-view information to improve 3D subject consistency.
Tri-Prompting combines multi-view subject conditioning with explicit camera and subject motion control to preserve identity under large view changes.

\vspace{-2mm}
\section{Method}
\vspace{-2mm}
\subsection{Overview}
Tri-Prompting consists of a video diffusion model guided by first-frame and multi-view reference images, and a motion control module driven by dual-conditioning 3D control cues, as shown in \cref{fig:method}. This framework jointly controls scene, multi-view subject, and motion in video generation. It is also designed to preserve multi-view appearance identity of the given subject, and achieve flexible disentangled motion control between foreground object and background scenes.

\begin{figure*}[t]
    \centering
    \includegraphics[width=0.7\linewidth]{Fig/method.pdf}
    \caption{\textbf{Overview of the proposed Tri-Prompting.} Tri-Prompting integrates video diffusion with first-frame/multi-view images and a dual-conditioning motion anchor to jointly control scene, subject and motion. We use two-stage training: first optimizing LoRA for scene and subject control, followed by ControlNet finetuning for motion control. The framework preserves multi-view identity while supporting decoupled motion control between the foreground and background. }
    \label{fig:method}
    \vspace{-3mm}
\end{figure*}

During inference, Tri-Prompting requires three inputs: (1) an image $I\in \mathbb{R}^{H \times W \times 3}$ with a text prompt that defines the scene layout and initial composition; (2) multi-view images (up to three) $\{S_i\}$ of the same single subject, where each $S_i \in \mathbb{R}^{H \times W \times 3}$, which define subject identity and unseen appearance; (3) a motion-control video $M \in \mathbb{R}^{T \times H \times W \times C}$ containing XYZ points for background/camera motion and low-resolution RGB proxies for foreground subject motion. This video can be generated from a reference video, pose-matrix transformations, or an offline control interface with keyboard. The output is a video $V \in \mathbb{R}^{T \times H \times W \times C}$.

\cref{method:stage1} introduces the proposed multi-view subject I2V model, and \cref{method:stage2} presents the ControlNet-based motion control module with dual 3D cues. Finally, \cref{method:inference} describes the inference pipeline and different practical workflows.

\subsection{Multi-view Subject-Image-to-Video}\label{method:stage1}
Tri-Prompting requires two stages of training to make the optimization more stable, effective, and trackable.
Stage 1 focuses on extending the single-view subject-to-video model into a joint image and multi-view subject-to-video model, establishing foundational control over both the background scene (via the first-frame image) and 3D-aware subject identity (via multi-view reference images).
Specifically, we encode the first-frame image and multi-view subject images with the base video diffusion's VAE encoder to get $z_I \in \mathbb{R}^{1 \times \frac{H}{sc} \times \frac{W}{sc} \times C} $ and $z_{S} \in \mathbb{R}^{k \times \frac{H}{sc} \times \frac{W}{sc} \times C}$, where $k$ is the number of multi-view subject reference images, $sc$ is the spatial compression factor, and $C$ is the latent channel number. We then prepend $z_I$ to the original noisy video latents $z_V \in \mathbb{R}^{\frac{T}{tc} \times \frac{H}{sc} \times \frac{W}{sc} \times C }$, where $tc$ is the temporal compression ratio of the VAE, and subsequently append $z_S$ to be the input token sequence $z_{seq}$ of DiT blocks:

\vspace{-2mm}
\begin{equation}
z_{seq} \leftarrow [ z_I , z_V , z_S ] \in \mathbb{R}^{(1 + \frac{T}{tc} + k) \times \frac{H}{sc} \times \frac{W}{sc} \times C}
\end{equation}

The order of the three latent components can be swapped if full self-attention is used in the video diffusion model. We apply LoRA~\cite{lora} on the attention and MLP blocks to condition on $z_I$ for first-frame generation and on $z_S$ for cross-view identity. During decoding, only $z_V$ is kept to generate the final video $V$.

\subsection{Dual-Conditioning Motion Control Module}\label{method:stage2}

The first stage already yields a video diffusion model that follows the first-frame scene and preserves identity across multiple views of the subject. In the second stage, we add explicit motion control over both the scene and the subject with dual control cues.

We first revisit two types of explicit control signals for motion generation: 3D tracking points and 2D RGB. 3D tracking points (XYZ trajectories) are effective control signals in previous works, such as DaS \cite{das}, but they mainly constrain visible and trackable surfaces. Because XYZ tracks are recovered from single-view videos, they capture only the geometry directly visible to the camera. For foreground subjects, rotation and self-occlusion can reveal newly visible regions that have no corresponding tracks, making tracking-point-only control unreliable for large viewpoint changes. Another potential control condition is a dense RGB pixel signal produced by optical flow, as in EPiC \cite{wang2025epic}. Dense RGB pixels tend to encourage the model to learn hole-filling behaviors like inpainting, yielding artifacts around hole regions.

To address these problems, we define a dual-conditioning signal: 3D tracking points for background and downsampled RGB proxies for foreground. Intuitively, XYZ tracking points are suitable for visible and trackable background surfaces under camera motion, while low-resolution RGB proxies encode coarse 3D-aware subject pose and spatial motion for foreground. These proxies intentionally avoid specifying fine appearance details, encouraging the model to rely on multi-view references and its generative prior to recover subject appearance and plausible subject-scene interactions.
Root-6DoF controls~\cite{fu2025trajmaster} require accurate object poses and canonical frames, which are difficult to obtain for diverse open-world subjects. Low-resolution RGB proxies can be constructed directly from videos or renders without object-specific pose annotations.

Specifically, we define the anchor motion control video as $M=\{M_{scene},M_{subject}\}$. For scene (background) control, we follow DaS to construct an XYZ tracking point video $M_{scene} \in \mathbb{R}^{T \times H \times W \times C}$. Each point's 3D coordinates (XYZ) are determined based on its position and depth in the first frame. These coordinates are then normalized to [0, 1] and converted to pseudo-RGB $c_i$. The color of the same tracking point does not change to keep the identity of the point. For subject (foreground) control, we use a low-resolution downsampled RGB point proxy $M_{subject} \in \mathbb{R}^{T \times H \times W \times C}$ obtained by downsampling the subject pixels into a fixed grid (e.g., $70 \times 70$) within the subject region. We composite these two conditions into a single anchor video in a spatially exclusive manner. This proxy provides sufficient cues of the camera pose as DaS \cite{das}, and provides stronger guidance on the object motion. Missing appearance details are complemented by the multi-view subject references via self-attention layers and by the base model prior.

We leverage this proxy and develop a ControlNet on top of the stage 1 base model. We copy the trained weights from the stage 1 model and add zero-initialized layers following the ControlNet architecture. The same video diffusion VAE encoder encodes $M$ to $z_M \in \mathbb{R}^{\frac{T}{tc} \times \frac{H}{sc} \times \frac{W}{sc} \times C}$. Similarly to \cref{method:stage1}, the input of the ControlNet DiT block is generated by concatenating $z_I$, $z_M$ and $z_S$. And only the $z_M$ of output will be used to update $z_V$.

\vspace{-2mm}
\begin{equation}
z_V \leftarrow z_V + s \cdot \mathrm{ControlNet}([ z_I , z_M , z_S ])[1:1 + \frac{T}{tc}]
\end{equation}

where $s$ represents the guidance scale of ControlNet. During the stage 2 finetuning, the weights of the base model are fixed, and only the ControlNet's weights are updated.

\subsection{Inference}\label{method:inference}

\subsubsection{Inference-time ControlNet Scale Schedule}\label{method:inference_aug}
During training, the low-resolution RGB condition is derived directly from ground truth video, ensuring perfect alignment with the target.
At inference time, user-specified 3D transformations provide coarse motion but may not contain natural micro-motions, such as leg lifts during walking, leading to rigid results.
Although diffusion priors can bring realism, they remain limited when over-constrained by these motion cues.

To address this issue, we introduce an optional ControlNet scale schedule to balance the trade-off between controllability and video realism. While the model is trained with a fixed ControlNet scale of 1.0, during inference we gradually reduce this influence to prevent over-constraining the generation. Specifically, for a sampling process with 50 denoising steps, we linearly anneal the scale during the first $N_{\text{decay}}$ steps and keep it constant thereafter:

\vspace{-2mm}
\begin{equation}
\text{s}(t) =
\begin{cases}
1 - \frac{t}{N_{\text{decay}}} (1 - s_{\min}) & \text{if } t \le N_{\text{decay}},\\[4pt]
s_{\min} & \text{otherwise.}
\end{cases}
\end{equation}

where $s_{\text{min}}$ is the final ControlNet scale after decay.

\subsubsection{Inference Workflow}\label{method:inference_workflow}

Tri-Prompting enables several new workflows that were previously difficult or impossible.
We introduce two novel workflows using an offline control interface and an inference pipeline that support identity-preserved motion control. The base model accepts text prompts for further flexibility.

Users can choose the scene (first frame image), select the character (multi-view subject), and drive the scene and subject motion.
We developed an offline control interface that allows users to drive motion via keyboard controls for 3D subject translation/rotation and camera pose.
Given subject reference images, we reconstruct a 3D subject using the Gaussian output of TRELLIS~\cite{trellis}, render downsampled 2D projections per frame, and paste it as the foreground RGB guidance.
For the background, we estimate the first-frame depth with DepthPro~\cite{depthpro} and convert camera pose into XYZ point trajectories.
These two spatially exclusive signals form the dual conditioning for motion control.

\noindent\textbf{Insertion of 3D Subjects into Scenes and Joint Control.}
Given a background scene image and a 3D subject, we first create a harmonized initial frame by inserting the subject’s initial 2D projection using an image editing model (e.g., Gemini~\cite{team2023gemini}, FluxKontext~\cite{batifol2025flux}, or Photoshop Generative Fill).
We then provide three representative subject views to the model.
Starting from this frame, users can control camera and subject motion independently or jointly:
(i) camera control is specified by transforming only the background XYZ points;
(ii) subject control is specified by applying translations/rotations to the reconstructed 3D subject.
The low-resolution foreground guidance steers motion rather than appearance, while multi-view references preserve identity and 3D-consistent details.

\noindent\textbf{Manipulation of 3D Subjects in a Scene and Joint Control.}
Given a scene image containing multiple subjects, we use it as the first frame, obtain a target mask via SAM~2~\cite{sam2}, and reconstruct the selected subject in 3D with SAM~3D~\cite{sam3d}.
We render representative multi-view references from the reconstructed 3D asset, optionally refined by an image editing model.
With the same dual-conditioning construction, users can control camera pose and subject motion to manipulate the scene with identity-preserved, motion-controlled generation.

\vspace{-4mm}
\section{Experiment}
\label{sec:experiment}
\vspace{-2mm}

\subsection{Training Data}

Training Tri-Prompting requires three main assets from videos ($V$): the first frame (I), three multi-view reference images ($\{S_i\}$) of the main subject in the scene, and a synthesized motion anchor video ($M$). To make the training efficient, we require the dataset videos to contain a single salient subject with extreme pose changes. The categories of the subjects better cover both rigid objects and non-rigid characters.
Therefore, we constructed 11k tuples $(I,\{S_i\},M,V)$ (total $<$7 hour-video), comprising 9.7k game videos from the OmniWorld-Game dataset~\cite{omniworld} and 1.3k real-world videos from the CO3D dataset~\cite{co3d}.
Despite this focused training, we observe qualitative generalization across scenes/subjects/poses and diverse styles (e.g., anime, film); see \cref{exp:application} for illustration.

Each tuple was constructed as follows:
(1) First frame ($I$): We took the first frame of each video as the scene image. (2) Multi-view subject images ($\{S_i\}$): For OmniWorld-Game, we manually identified and cropped three views of the same subject across frames. For the CO3D dataset, which is object-centric with rotating views, we directly sampled frames around the object as the multi-view subject references $\{S_i\}$. (3) Motion control videos ($M$). Each dataset provides the masks of the salient subjects. We separate the background and foreground subject to build the dual-conditioning motion control videos. For the background, we follow DaS to extract an XYZ tracking video: 3D points are tracked by SpatialTracker~\cite{SpatialTracker} across frames and normalized. For the subject, we construct low-resolution RGB points by downsampling the whole video from $832 \times 480$ to a fixed grid ($70{\times}70$), and keeping only grids within the subject mask.

\subsection{Experiment Details and Settings}

We use a two-stage training strategy in Tri-Prompting. In stage 1, we fine-tune the Phantom S2V 14B backbone \cite{phantom} with LoRA (rank 64) on attention and MLP blocks. Training utilizes AdamW (lr $1\!\times\!10^{-4}$), with a batch size of 8 for 2500 steps. This stage takes 20 hours on 8 A100 GPUs. In stage 2, we initialize a ControlNet by copying the first 18 layers from the first stage checkpoint. Then we freeze the base diffusion model and finetune the added ControlNet modules.
Control injections are zero-initialized (zero-conv). Training also utilizes AdamW (lr $1\!\times\!10^{-4}$), with a batch size of 32 for 2074 steps. The second stage takes 28 hours using 32 A100 GPUs.

\subsection{Comparison with Baselines}
We compare with DaS~\cite{das} for video reconstruction, and Phantom~\cite{phantom} and OmniVCus~\cite{cai2025omnivcus} for subject identity, video quality, and 3D consistency.
Additional comparisons with MotionBooth are provided in \cref{supp:motionbooth}.

\begin{table*}[t]
    \centering
    \caption{\textbf{Comparison with specialized baselines.} (a) Video reconstruction on DAVIS. (b) Video quality, multi-view identity, and 3D consistency on our evaluation set; geometry scores use 84 videos successfully reconstructed across all settings.}
    \label{tab:main_results}
    \begin{subtable}{\textwidth}
        \centering
        \caption{Comparison with DaS on DAVIS}
        \label{tab:result_a}
        \begin{adjustbox}{max width=\textwidth}
        \begin{tabular}{lllcccc}
            \toprule
             & \textbf{Motion control} & \textbf{Base Model} & \textbf{PSNR$\uparrow$} & \textbf{SSIM$\uparrow$} & \textbf{LPIPS$\downarrow$} \\
            \midrule
            DaS \cite{das} & XYZ tracking & CogVideoX (5B) \cite{cogvideox} & \textit{16.4916} & \textbf{0.4123} & 0.2725 \\
            Ours & XYZ tracking & Phantom (14B) \cite{phantom} & 15.9306 & 0.3708 & 0.2786 \\
            Ours & XYZ tracking \& RGB  & Phantom (14B) \cite{phantom} & \textbf{16.5130} & \textit{0.4017} & \textbf{0.2395} \\
            \bottomrule
        \end{tabular}
        \end{adjustbox}
        \vspace{1mm} %
    \end{subtable}

    \begin{subtable}{\textwidth}
        \centering
        \caption{Comparison with Phantom and OmniVCus on VBench and Multi-view Consistency}
        \label{tab:result_b}
        \begin{adjustbox}{max width=\textwidth}
        \begin{tabular}{l l cccc cc c}
            \toprule
            \multirow{2}{*}{Method} & \multirow{2}{*}{\shortstack{Motion\\Anchor}}& \multicolumn{4}{c}{VBench Quality $\uparrow$} & \multicolumn{2}{c}{MV Subj. ID $\uparrow$} & \multicolumn{1}{c}{3D Consis. $\downarrow$} \\
            \cmidrule(lr){3-6} \cmidrule(lr){7-8} \cmidrule(lr){9-9}
            && Aes. & Mot. & Dyn. & Img. & $\mathcal{I}_{\text{dino}}$ & $\mathcal{I}_{\text{clip}}$ & $E_{\text{align}}$ \\
            \midrule
            Phantom (3-view) & No & 0.655 & 0.988 & 0.742 & 0.709 & 0.723 & 0.855 & 0.034 \\
            OmniVCus (3-view montage) & Mask & 0.618 & 0.987 & 0.613 & 0.703 & 0.703 & 0.846 & 0.044 \\
            \textbf{Ours stage1 (3-view)} & No & 0.660 & 0.987 & 0.682 & 0.705 & 0.732 & 0.861 & 0.027 \\
            \textbf{Ours stage2 (3-view)} & Yes & \textbf{0.665} & \textbf{0.989} & \textbf{0.903} & \textbf{0.715} & \textbf{0.746} & \textbf{0.862} & \textbf{0.025} \\
            \bottomrule
        \end{tabular}
        \end{adjustbox}
        \vspace{1mm}
    \end{subtable}

    \vspace{-3mm}
\end{table*}

\begin{figure*}[t]
    \centering
    \includegraphics[width=1.\linewidth]{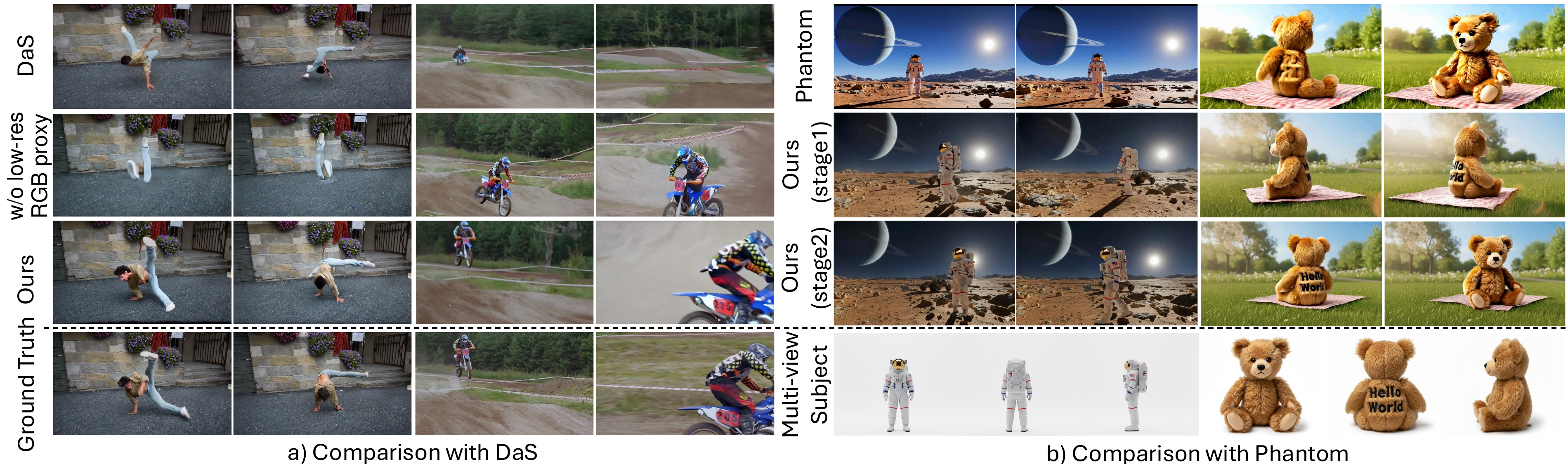}
    \caption{\textbf{Comparison with DaS and Phantom.} Tri-Prompting improves motion robustness over tracking-only control and preserves multi-view subject identity under large pose changes.}
    \label{fig:comparison}
    \vspace{-5mm}
\end{figure*}

\noindent\textbf{Comparison with DaS}: Following DaS, we evaluate the video reconstruction performance under multi-conditioning using PSNR, SSIM, and LPIPS, demonstrating motion controllability of our model.
We evaluate on 50 videos from the DAVIS dataset~\cite{perazzi2016benchmark}, covering diverse motion patterns, scenes, and subjects.
We use the official DaS checkpoint to test on DAVIS, as shown in \cref{tab:main_results}(a). %

DaS reconstructs the video from the first frame and its XYZ tracking points. Tri-Prompting additionally uses low-resolution foreground RGB and multi-view subject references, yielding higher PSNR and lower LPIPS. Its lower SSIM and weaker XYZ-only result may reflect Phantom's focus on reference-conditioned identity compared with CogVideoX's first-frame I2V focus. We use Phantom for its subject-reference interface and isolate the RGB contribution through same-backbone ablations (\cref{ablation}).

As illustrated in \cref{fig:comparison}, our method maintains robust motion alignment under extreme human actions where tracking-point-only methods fail and hallucinate as points disappear. Crucially, multi-view references enable the accurate recovery of intricate details like text and patterns that are occluded in the first frame—details that baseline models must otherwise hallucinate.

\noindent\textbf{Comparison with Phantom}: We compare against Phantom across three dimensions: video quality, multi-view subject similarity, and 3D consistency. We define 106 test cases from scene, subject, and motion inputs before generation and evaluate all 318 outputs (3 random seeds). For each case, both models share the scene, 3-view subject images, and prompt; ours can also take additional motion anchors.

For video quality, we use four metrics from VBench~\cite{vbench}: aesthetic quality, dynamic degree, imaging quality, and motion smoothness.
For multi-view ID, $\mathcal{I}_{dino/clip} = \text{mean}_i \max_j \text{sim}(\mathbf{f}_i, \mathbf{s}_j)$ compares DINO~\cite{dinov2}/CLIP~\cite{clip} features of segmented subject regions \cite{sam2} in generated frames $\mathbf{f}$ with subject images $\mathbf{s}$.
For 3D consistency, we reconstruct 3D points $\mathcal{P}_g$ and $\mathcal{P}_s$ from subject region of generated videos and subject images using reconstruction model $\pi^3$~\cite{pi} and measure 3D alignment error $E_{\text{align}} = \text{mean}_{\mathbf{x} \in \mathcal{P}_g} \min_{\mathbf{y} \in \mathcal{P}_s} \|\mathbf{x} - \mathbf{y}\|_2$. In \cref{tab:main_results}(b), we evaluate rigid subjects (\textit{e.g.}, cars, chairs, and bags) on 84 outputs successfully reconstructed across all evaluated settings; protocol details are in \cref{supp:omnivcus}.
Results covering all subjects are reported in \cref{supp:3d_all}.

As shown in \cref{tab:main_results}(b), Tri-Prompting outperforms Phantom across all metrics, reducing $E_{\text{align}}$ from 0.034 to 0.025.
This result underscores a practical advantage: while Phantom mainly enforces 2D appearance consistency, our method better preserves 3D integrity across frames while following motion control.
Qualitative results in \cref{fig:comparison} further highlight these differences; Phantom exhibits structural distortions during multi-view fusion—for instance, an astronaut’s body may face backward while moving forward, and a rotating teddy bear suffers from warping and loss of shape. In contrast, Tri-Prompting maintains more consistent identity and 3D structure under such large pose changes.

\noindent\textbf{Comparison with OmniVCus.}
We evaluate the released OmniVCus checkpoint on the same 106 cases and three seeds, using text, foreground-mask videos, and a fixed montage of the three subject views. Tri-Prompting improves video quality, multi-view identity, and 3D consistency over OmniVCus (\cref{tab:main_results}(b)). Single-reference results and input details are provided in \cref{supp:omnivcus}.

\begin{figure*}[t]
    \centering
    \includegraphics[width=0.8\linewidth]{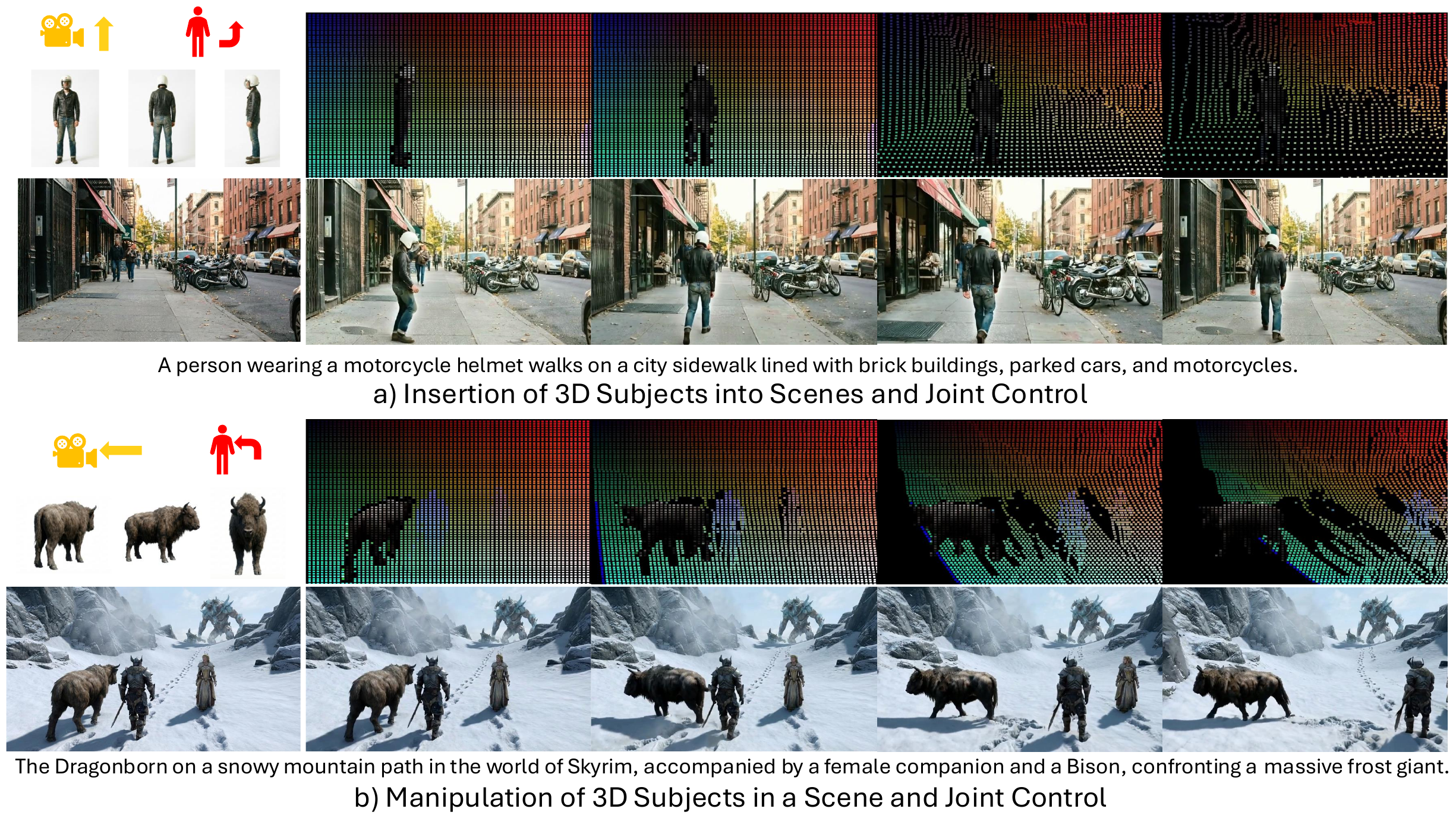}
    \caption{\textbf{Applications.} Tri-Prompting supports 3D-aware subject insertion and in-scene manipulation with camera/subject motion control. The multi-view identity is also preserved, providing strong 3D consistency. Please see more video results in appendix.}
    \label{fig:result}
    \vspace{-3mm}
\end{figure*}

\vspace{-2mm}
\subsection{Applications} \label{exp:application}

Some qualitative results are shown in \cref{fig:result}, highlighting both (i) inserting multi-view subjects into a scene and (ii) manipulating existing subjects within a scene. In both applications, the resulting foreground-background interactions remain visually coherent. For instance, under camera-pose changes, the cowboy and Wukong (monkey) exhibit plausible non-rigid motion (e.g., walking) consistent with the scene dynamics. Additionally, the multi-view appearance from the reference subject images is preserved, providing strong identity consistency.

Tri-Prompting provides robust, generalizable control over scene, subject, and motion.
Although fine-tuned on a mix of game~\cite{omniworld} and common subjects videos~\cite{co3d}, we observe generalization to scenes/subjects/motion and diverse styles (anime/movie/real-world) in \cref{fig:result} and Supp., demonstrating our model works well on out-of-distribution samples.

\vspace{-1mm}
\subsection{Ablation Studies}
\label{ablation}
We ablate dual-conditioning, training stages, reference views, ControlNet scale scheduling, and RGB resolution. Public 360Motion results in \cref{tab:scale_schedule} are analyzed in \cref{exp:generalization}.

\noindent\textbf{Dual-conditioning vs Tracking Points Only.}
To evaluate the benefits of RGB conditioning, we compare different motion control signals on the same Phantom base model, as shown in the second and third rows of \cref{tab:main_results}(a). Dual-conditioning achieves superior PSNR, SSIM, and LPIPS compared to using XYZ tracking alone. This demonstrates that incorporating RGB proxies provides more reliable guidance for extreme poses and 3D motion. In contrast, XYZ tracking alone often fails to propagate complex dynamics due to sparse or missing points.

\noindent\textbf{Effect of Training Stages.}
We observe progressive performance gains across our two-stage training strategy (\cref{tab:main_results}(b)). While Stage 1 establishes a robust foundation for multi-view identity, the introduction of dual-conditioning motion control in Stage 2 serves as more than a functional addition—it acts as a guiding anchor that further bolsters overall video quality, multi-view identity, and 3D consistency. This progression suggests that identity-focused training and motion-guided refinement work together to improve appearance preservation and motion control.

\begin{figure}[htbp]
    \centering
    \includegraphics[width=0.8\linewidth]{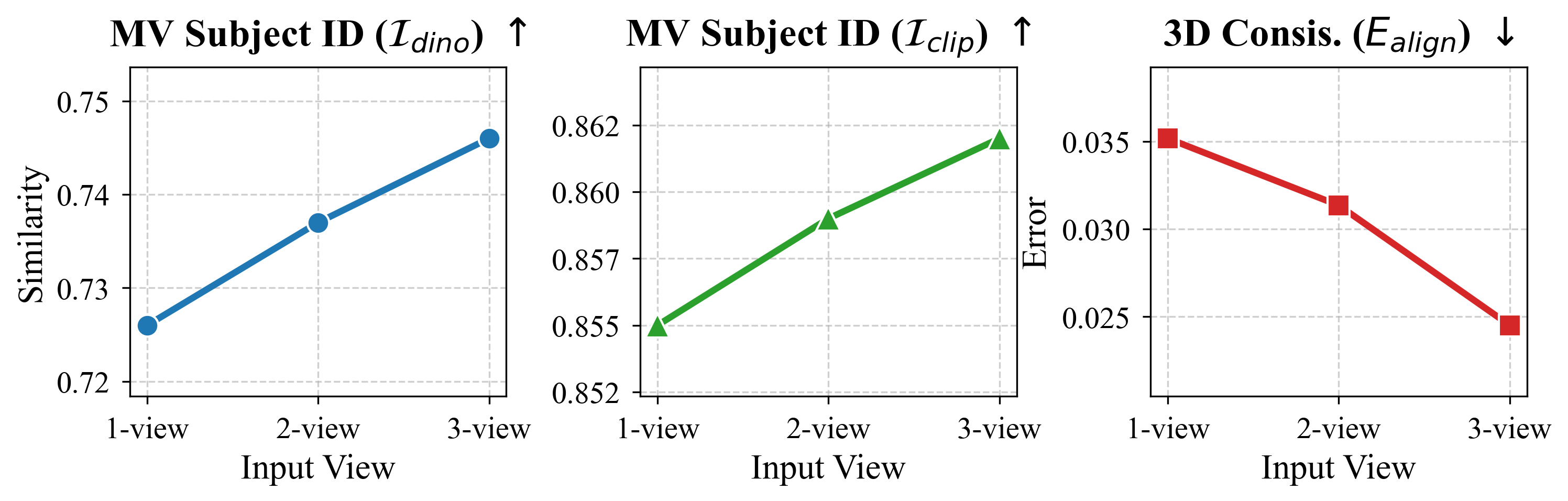}
    \caption{\textbf{Effect of multi-view subject images.} More views improve multi-view ID and 3D consistency, reducing ambiguity during pose changes.}
    \label{fig:views}
\end{figure}

\noindent\textbf{Effect of Multi-view Subject Images.}
We finetune and infer models with 1/2/3 views to test the effects of multi-view subject views. As shown in \cref{fig:views}, all metrics improve with more views. Providing 3 views reduces the 3D alignment error ($E_{align}$) from 0.035 to 0.025 and consistently boosts identity similarity. Single-reference input is supported, while additional views reduce appearance ambiguity during pose changes.

\begin{figure}[htbp]
    \centering
    \includegraphics[width=0.75\linewidth]{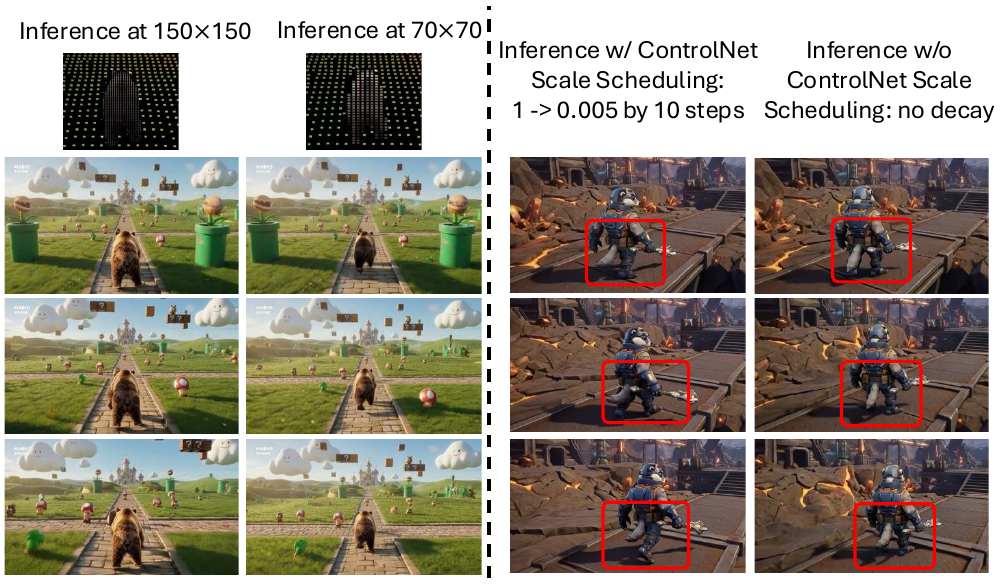}
    \caption{\textbf{Ablation study.} Left: Low-resolution RGB enables plausible subject-scene interactions. Right: ControlNet scale decay reduces rigid motion.}
    \label{fig:ablation}
\end{figure}

\noindent\textbf{Effect of ControlNet Scale Scheduling.}
ControlNet scale scheduling balances controllability and visual realism. We compare constant scale 1.0 with decay from 1.0 to 0.005 over the first 10 of 50 denoising steps, using the same settings for all videos on both datasets. In \cref{fig:ablation}, decay relaxes the rigid translation cue and enables a more natural leg-lift gait, but may reduce motion controllability. On both datasets, decay improves multi-view identity with lower Bbox/Mask IoU (\cref{tab:scale_schedule}).

\noindent\textbf{Low-res RGB vs. High-res RGB.}
As shown in \cref{fig:ablation}, the low-resolution RGB condition provides greater flexibility for plausible subject-scene interactions, while higher-resolution RGB better preserves the rigid properties of objects when needed.
Users can also adjust the RGB resolution during inference to control the degree of rigidity in the object’s motion.

\vspace{-1mm}
\subsection{Generalization Studies}
\label{exp:generalization}
Our 106-case evaluation set contains no training cases and tests Tri-Prompting's out-of-distribution (OOD) generalization to unseen scene/subject/pose combinations across anime, movie, game, and real-world styles. We further use released videos from 360Motion~\cite{fu2025trajmaster}, the dataset used by 3DTrajMaster (35 single-entity cases from the held-out C12 split, three seeds per case). 360Motion is in-domain for 3DTrajMaster but unseen during our training. We compute Bbox/Mask IoU against these reference videos using our evaluation protocol (\cref{supp:360motion}). On 360Motion, constant scale yields higher Bbox/Mask IoU than 3DTrajMaster (0.632/0.510 vs.\ 0.590/0.435), while scale decay improves DINO identity similarity over Phantom (0.654 vs.\ 0.639) with comparable CLIP similarity (\cref{tab:scale_schedule}). This external evaluation further supports generalization of motion control and identity preservation.

\begin{table}[htbp]
    \centering
    \caption{\textbf{Generalization and effects of ControlNet scale scheduling.} Motion adherence and identity on our evaluation set (318 outputs) and public 360Motion (105 outputs). Scale decay improves identity while reducing motion adherence.}
    \label{tab:scale_schedule}
    \small
    \setlength{\tabcolsep}{3pt}
    \begin{adjustbox}{max width=\textwidth}
    \begin{tabular}{@{}lcccc@{\hspace{10pt}}cccc@{}}
        \toprule
        \multirow{2}{*}{Method} & \multicolumn{4}{c}{Our evaluation set} & \multicolumn{4}{c}{Public 360Motion} \\
        \cmidrule(lr){2-5} \cmidrule(lr){6-9}
         & Bbox IoU$\uparrow$ & Mask IoU$\uparrow$ & $\mathcal{I}_{\text{dino}}\!\uparrow$ & $\mathcal{I}_{\text{clip}}\!\uparrow$ & Bbox IoU$\uparrow$ & Mask IoU$\uparrow$ & $\mathcal{I}_{\text{dino}}\!\uparrow$ & $\mathcal{I}_{\text{clip}}\!\uparrow$ \\
        \midrule
        Phantom & --- & --- & 0.723 & 0.855 & --- & --- & 0.639 & 0.874 \\
        3DTrajMaster & 0.652 & 0.540 & --- & --- & 0.590 & 0.435 & --- & --- \\
        \midrule
        Ours, constant scale & \textbf{0.861} & \textbf{0.790} & 0.733 & 0.858 & \textbf{0.632} & \textbf{0.510} & 0.627 & 0.864 \\
        Ours, scale decay & 0.775 & 0.684 & \textbf{0.746} & \textbf{0.862} & 0.535 & 0.411 & \textbf{0.654} & \textbf{0.875} \\
        \bottomrule
    \end{tabular}
    \end{adjustbox}
\end{table}

\vspace{-2mm}
\subsection{Limitation}
\vspace{-1mm}
\label{limitation}

Highly symmetric multi-view references can occasionally cause brief identity inconsistencies, such as transient flips, which can be alleviated by asymmetric references or stronger motion constraints.
The pipeline also depends on external segmentation, reconstruction, and view-generation components; severe errors in their outputs can degrade generation. Large camera rotations can weaken background control when the XYZ points visible in the first frame leave the view.
Our 14B model targets offline generation, taking about 3.85 minutes on eight A100-80GB GPUs for a 49-frame, $832\times480$ video with 50 denoising steps (\cref{supp:cost}). Video diffusion acceleration could reduce this latency. The current method controls one primary subject and does not support complex multi-subject control.

\vspace{-2mm}
\section{Conclusion}
\vspace{-1mm}
We present Tri-Prompting, a video diffusion framework that integrates scene, subject, and motion prompts. By combining multi-view references with a dual-conditioning module—XYZ coordinates for backgrounds and low-resolution RGB proxies for foregrounds—we recover detailed, 3D-consistent subjects from coarse motion cues.
In evaluations, Tri-Prompting achieves competitive or improved performance against specialized baselines: it obtains competitive PSNR/SSIM and improved perceptual similarity compared to DaS, and outperforms Phantom in multi-view subject-to-video generation.
To support representative applications such as insertion and manipulation of 3D-aware subjects in a scene, we further develop a control interface to construct the motion video. Tri-Prompting supports scene/subject motion control across diverse scenes and styles.
In the future, we plan to explore more efficient designs to reduce inference time, moving toward an interactive
experience that can better support world-simulator-style applications.

\renewcommand{\acksection}{\section*{Acknowledgments}}
\begin{ack}
This work was partially supported by the university's Goergen Institute for Data Science and Artificial Intelligence.
\end{ack}

\bibliographystyle{plainnat}
\bibliography{main}

@inproceedings{das,
  title={Diffusion as shader: 3d-aware video diffusion for versatile video generation control},
  author={Gu, Zekai and Yan, Rui and Lu, Jiahao and Li, Peng and Dou, Zhiyang and Si, Chenyang and Dong, Zhen and Liu, Qifeng and Lin, Cheng and Liu, Ziwei and others},
  booktitle={SIGGRAPH},
  year={2025}
}

@article{phantom,
  title={Phantom: Subject-consistent video generation via cross-modal alignment},
  author={Liu, Lijie and Ma, Tianxiang and Li, Bingchuan and Chen, Zhuowei and Liu, Jiawei and Li, Gen and Zhou, Siyu and He, Qian and Wu, Xinglong},
  journal={ICCV},
  year={2025}
}

@article{wan,
      title={Wan: Open and Advanced Large-Scale Video Generative Models},
      author={Team Wan and Ang Wang and Baole Ai and Bin Wen and Chaojie Mao and Chen-Wei Xie and Di Chen and Feiwu Yu and Haiming Zhao and Jianxiao Yang and Jianyuan Zeng and Jiayu Wang and Jingfeng Zhang and Jingren Zhou and Jinkai Wang and Jixuan Chen and Kai Zhu and Kang Zhao and Keyu Yan and Lianghua Huang and Mengyang Feng and Ningyi Zhang and Pandeng Li and Pingyu Wu and Ruihang Chu and Ruili Feng and Shiwei Zhang and Siyang Sun and Tao Fang and Tianxing Wang and Tianyi Gui and Tingyu Weng and Tong Shen and Wei Lin and Wei Wang and Wei Wang and Wenmeng Zhou and Wente Wang and Wenting Shen and Wenyuan Yu and Xianzhong Shi and Xiaoming Huang and Xin Xu and Yan Kou and Yangyu Lv and Yifei Li and Yijing Liu and Yiming Wang and Yingya Zhang and Yitong Huang and Yong Li and You Wu and Yu Liu and Yulin Pan and Yun Zheng and Yuntao Hong and Yupeng Shi and Yutong Feng and Zeyinzi Jiang and Zhen Han and Zhi-Fan Wu and Ziyu Liu},
      journal = {arXiv preprint arXiv:2503.20314},
      year={2025}
}

@article{cogvideox,
  title={Cogvideox: Text-to-video diffusion models with an expert transformer},
  author={Yang, Zhuoyi and Teng, Jiayan and Zheng, Wendi and Ding, Ming and Huang, Shiyu and Xu, Jiazheng and Yang, Yuanming and Hong, Wenyi and Zhang, Xiaohan and Feng, Guanyu and others},
  journal={ICLR},
  year={2025}
}

@inproceedings{motionprompting,
  title={Motion prompting: Controlling video generation with motion trajectories},
  author={Geng, Daniel and Herrmann, Charles and Hur, Junhwa and Cole, Forrester and Zhang, Serena and Pfaff, Tobias and Lopez-Guevara, Tatiana and Aytar, Yusuf and Rubinstein, Michael and Sun, Chen and others},
  booktitle={CVPR},
  year={2025}
}

@article{omniworld,
  title={Omniworld: A multi-domain and multi-modal dataset for 4d world modeling},
  author={Zhou, Yang and Wang, Yifan and Zhou, Jianjun and Chang, Wenzheng and Guo, Haoyu and Li, Zizun and Ma, Kaijing and Li, Xinyue and Wang, Yating and others},
  journal={arXiv preprint arXiv:2509.12201},
  year={2025}
}

@inproceedings{lora,
title={Lo{RA}: Low-Rank Adaptation of Large Language Models},
author={Edward J Hu and Yelong Shen and Phillip Wallis and Zeyuan Allen-Zhu and Yuanzhi Li and Shean Wang and Lu Wang and Weizhu Chen},
booktitle={ICLR},
year={2022}
}

@inproceedings{co3d,
	Author = {Reizenstein, Jeremy and Shapovalov, Roman and Henzler, Philipp and Sbordone, Luca and Labatut, Patrick and Novotny, David},
	Booktitle = {ICCV},
	Title = {Common Objects in 3D: Large-Scale Learning and Evaluation of Real-life 3D Category Reconstruction},
	Year = {2021},
}

@inproceedings{SpatialTracker,
    title={SpatialTracker: Tracking Any 2D Pixels in 3D Space},
    author={Xiao, Yuxi and Wang, Qianqian and Zhang, Shangzhan and Xue, Nan and Peng, Sida and Shen, Yujun and Zhou, Xiaowei},
    booktitle={CVPR},
    year={2024}
}

@article{wang2025epic,
  title={EPiC: Efficient Video Camera Control Learning with Precise Anchor-Video Guidance},
  author={Wang, Zun and Cho, Jaemin and Li, Jialu and Lin, Han and Yoon, Jaehong and Zhang, Yue and Bansal, Mohit},
  journal={arXiv preprint arXiv:2505.21876},
  year={2025}
}

@inproceedings{depthpro,
  author     = {Aleksei Bochkovskii and Ama\"{e}l Delaunoy and Hugo Germain and Marcel Santos and
               Yichao Zhou and Stephan R. Richter and Vladlen Koltun},
  title      = {Depth Pro: Sharp Monocular Metric Depth in Less Than a Second},
  booktitle  = {ICLR},
  year       = {2025},
}

@article{trellis,
    title   = {Structured 3D Latents for Scalable and Versatile 3D Generation},
    author  = {Xiang, Jianfeng and Lv, Zelong and Xu, Sicheng and Deng, Yu and Wang, Ruicheng and Zhang, Bowen and Chen, Dong and Tong, Xin and Yang, Jiaolong},
    journal = {arXiv preprint arXiv:2412.01506},
    year    = {2024}
}

@inproceedings{ma2024follow,
  title={Follow your pose: Pose-guided text-to-video generation using pose-free videos},
  author={Ma, Yue and He, Yingqing and Cun, Xiaodong and Wang, Xintao and Chen, Siran and Li, Xiu and Chen, Qifeng},
  booktitle={AAAI},
  volume={38},
  number={5},
  pages={4117--4125},
  year={2024}
}

@article{xing2024make,
  title={Make-your-video: Customized video generation using textual and structural guidance},
  author={Xing, Jinbo and Xia, Menghan and Liu, Yuxin and Zhang, Yuechen and Zhang, Yong and He, Yingqing and Liu, Hanyuan and Chen, Haoxin and Cun, Xiaodong and Wang, Xintao and others},
  journal={IEEE Transactions on Visualization and Computer Graphics},
  volume={31},
  number={2},
  pages={1526--1541},
  year={2024},
  publisher={IEEE}
}

@article{xing2024tooncrafter,
  title={Tooncrafter: Generative cartoon interpolation},
  author={Xing, Jinbo and Liu, Hanyuan and Xia, Menghan and Zhang, Yong and Wang, Xintao and Shan, Ying and Wong, Tien-Tsin},
  journal={TOG},
  volume={43},
  number={6},
  pages={1--11},
  year={2024},
  publisher={ACM New York, NY, USA}
}

@inproceedings{wang2024drivedreamer,
  title={Drivedreamer: Towards real-world-drive world models for autonomous driving},
  author={Wang, Xiaofeng and Zhu, Zheng and Huang, Guan and Chen, Xinze and Zhu, Jiagang and Lu, Jiwen},
  booktitle={ECCV},
  pages={55--72},
  year={2024},
  organization={Springer}
}

@inproceedings{chen2025echomimic,
  title={Echomimic: Lifelike audio-driven portrait animations through editable landmark conditions},
  author={Chen, Zhiyuan and Cao, Jiajiong and Chen, Zhiquan and Li, Yuming and Ma, Chenguang},
  booktitle={AAAI},
  volume={39},
  number={3},
  pages={2403--2410},
  year={2025}
}

@inproceedings{shi2024motion,
  title={Motion-i2v: Consistent and controllable image-to-video generation with explicit motion modeling},
  author={Shi, Xiaoyu and Huang, Zhaoyang and Wang, Fu-Yun and Bian, Weikang and Li, Dasong and Zhang, Yi and Zhang, Manyuan and Cheung, Ka Chun and See, Simon and Qin, Hongwei and others},
  booktitle={SIGGRAPH},
  pages={1--11},
  year={2024}
}

@article{wu2024motionbooth,
  title={Motionbooth: Motion-aware customized text-to-video generation},
  author={Wu, Jianzong and Li, Xiangtai and Zeng, Yanhong and Zhang, Jiangning and Zhou, Qianyu and Li, Yining and Tong, Yunhai and Chen, Kai},
  journal={NeurIPS},
  volume={37},
  pages={34322--34348},
  year={2024}
}

@article{yu2024viewcrafter,
  title={Viewcrafter: Taming video diffusion models for high-fidelity novel view synthesis},
  author={Yu, Wangbo and Xing, Jinbo and Yuan, Li and Hu, Wenbo and Li, Xiaoyu and Huang, Zhipeng and Gao, Xiangjun and Wong, Tien-Tsin and Shan, Ying and Tian, Yonghong},
  journal={arXiv preprint arXiv:2409.02048},
  year={2024}
}

@inproceedings{yang2024direct,
  title={Direct-a-video: Customized video generation with user-directed camera movement and object motion},
  author={Yang, Shiyuan and Hou, Liang and Huang, Haibin and Ma, Chongyang and Wan, Pengfei and Zhang, Di and Chen, Xiaodong and Liao, Jing},
  booktitle={SIGGRAPH},
  pages={1--12},
  year={2024}
}

@inproceedings{jiang2024videobooth,
  title={Videobooth: Diffusion-based video generation with image prompts},
  author={Jiang, Yuming and Wu, Tianxing and Yang, Shuai and Si, Chenyang and Lin, Dahua and Qiao, Yu and Loy, Chen Change and Liu, Ziwei},
  booktitle={CVPR},
  pages={6689--6700},
  year={2024}
}

@inproceedings{zhuang2024vlogger,
  title={Vlogger: Make your dream a vlog},
  author={Zhuang, Shaobin and Li, Kunchang and Chen, Xinyuan and Wang, Yaohui and Liu, Ziwei and Qiao, Yu and Wang, Yali},
  booktitle={CVPR},
  pages={8806--8817},
  year={2024}
}

@article{fei2025skyreels,
  title={Skyreels-a2: Compose anything in video diffusion transformers},
  author={Fei, Zhengcong and Li, Debang and Qiu, Di and Wang, Jiahua and Dou, Yikun and Wang, Rui and Xu, Jingtao and Fan, Mingyuan and Chen, Guibin and Li, Yang and others},
  journal={arXiv preprint arXiv:2504.02436},
  year={2025}
}

@article{cai2025omnivcus,
  title={OmniVCus: Feedforward Subject-driven Video Customization with Multimodal Control Conditions},
  author={Cai, Yuanhao and Zhang, He and Chen, Xi and Xing, Jinbo and Hu, Yiwei and Zhou, Yuqian and Zhang, Kai and Zhang, Zhifei and Kim, Soo Ye and Wang, Tianyu and others},
  journal={arXiv preprint arXiv:2506.23361},
  year={2025}
}

@article{ju2025editverse,
  title={EditVerse: Unifying Image and Video Editing and Generation with In-Context Learning},
  author={Ju, Xuan and Wang, Tianyu and Zhou, Yuqian and Zhang, He and Liu, Qing and Zhao, Nanxuan and Zhang, Zhifei and Li, Yijun and Cai, Yuanhao and Liu, Shaoteng and others},
  journal={arXiv preprint arXiv:2509.20360},
  year={2025}
}

@article{he2022latent,
  title={Latent video diffusion models for high-fidelity long video generation},
  author={He, Yingqing and Yang, Tianyu and Zhang, Yong and Shan, Ying and Chen, Qifeng},
  journal={arXiv preprint arXiv:2211.13221},
  year={2022}
}

@inproceedings{wu2023tune,
  title={Tune-a-video: One-shot tuning of image diffusion models for text-to-video generation},
  author={Wu, Jay Zhangjie and Ge, Yixiao and Wang, Xintao and Lei, Stan Weixian and Gu, Yuchao and Shi, Yufei and Hsu, Wynne and Shan, Ying and Qie, Xiaohu and Shou, Mike Zheng},
  booktitle={ICCV},
  pages={7623--7633},
  year={2023}
}

@article{guo2023animatediff,
  title={Animatediff: Animate your personalized text-to-image diffusion models without specific tuning},
  author={Guo, Yuwei and Yang, Ceyuan and Rao, Anyi and Liang, Zhengyang and Wang, Yaohui and Qiao, Yu and Agrawala, Maneesh and Lin, Dahua and Dai, Bo},
  journal={arXiv preprint arXiv:2307.04725},
  year={2023}
}

@article{blattmann2023stable,
  title={Stable video diffusion: Scaling latent video diffusion models to large datasets},
  author={Blattmann, Andreas and Dockhorn, Tim and Kulal, Sumith and Mendelevitch, Daniel and Kilian, Maciej and Lorenz, Dominik and Levi, Yam and English, Zion and Voleti, Vikram and Letts, Adam and others},
  journal={arXiv preprint arXiv:2311.15127},
  year={2023}
}

@article{brooks2024video,
  title={Video generation models as world simulators},
  author={Brooks, Tim and Peebles, Bill and Holmes, Connor and DePue, Will and Guo, Yufei and Jing, Li and Schnurr, David and Taylor, Joe and Luhman, Troy and Luhman, Eric and others},
  journal={OpenAI Blog},
  volume={1},
  number={8},
  pages={1},
  year={2024}
}

@article{kong2024hunyuanvideo,
  title={Hunyuanvideo: A systematic framework for large video generative models},
  author={Kong, Weijie and Tian, Qi and Zhang, Zijian and Min, Rox and Dai, Zuozhuo and Zhou, Jin and Xiong, Jiangfeng and Li, Xin and Wu, Bo and Zhang, Jianwei and others},
  journal={arXiv preprint arXiv:2412.03603},
  year={2024}
}

@article{ma2025step,
  title={Step-video-t2v technical report: The practice, challenges, and future of video foundation model},
  author={Ma, Guoqing and Huang, Haoyang and Yan, Kun and Chen, Liangyu and Duan, Nan and Yin, Shengming and Wan, Changyi and Ming, Ranchen and Song, Xiaoniu and Chen, Xing and others},
  journal={arXiv preprint arXiv:2502.10248},
  year={2025}
}

@article{polyak2024movie,
  title={Movie gen: A cast of media foundation models},
  author={Polyak, Adam and Zohar, Amit and Brown, Andrew and Tjandra, Andros and Sinha, Animesh and Lee, Ann and Vyas, Apoorv and Shi, Bowen and Ma, Chih-Yao and Chuang, Ching-Yao and others},
  journal={arXiv preprint arXiv:2410.13720},
  year={2024}
}

@article{team2023gemini,
  title={Gemini: a family of highly capable multimodal models},
  author={Team, Gemini and Anil, Rohan and Borgeaud, Sebastian and Alayrac, Jean-Baptiste and Yu, Jiahui and Soricut, Radu and Schalkwyk, Johan and Dai, Andrew M and Hauth, Anja and Millican, Katie and others},
  journal={arXiv preprint arXiv:2312.11805},
  year={2023}
}

@article{batifol2025flux,
  title={FLUX. 1 Kontext: Flow Matching for In-Context Image Generation and Editing in Latent Space},
  author={Batifol, Stephen and Blattmann, Andreas and Boesel, Frederic and Consul, Saksham and Diagne, Cyril and Dockhorn, Tim and English, Jack and English, Zion and Esser, Patrick and Kulal, Sumith and others},
  journal={arXiv e-prints},
  pages={arXiv--2506},
  year={2025}
}

@inproceedings{perazzi2016benchmark,
  title={A benchmark dataset and evaluation methodology for video object segmentation},
  author={Perazzi, Federico and Pont-Tuset, Jordi and McWilliams, Brian and Van Gool, Luc and Gross, Markus and Sorkine-Hornung, Alexander},
  booktitle={CVPR},
  pages={724--732},
  year={2016}
}

@article{sam2,
  title={Sam 2: Segment anything in images and videos},
  author={Ravi, Nikhila and Gabeur, Valentin and Hu, Yuan-Ting and Hu, Ronghang and Ryali, Chaitanya and Ma, Tengyu and Khedr, Haitham and R{\"a}dle, Roman and Rolland, Chloe and Gustafson, Laura and others},
  journal={arXiv preprint arXiv:2408.00714},
  year={2024}
}

@article{sam3d,
  title={Sam 3d: 3dfy anything in images},
  author={Chen, Xingyu and Chu, Fu-Jen and Gleize, Pierre and Liang, Kevin J and Sax, Alexander and Tang, Hao and Wang, Weiyao and Guo, Michelle and Hardin, Thibaut and Li, Xiang and others},
  journal={arXiv preprint arXiv:2511.16624},
  year={2025}
}

@InProceedings{vbench,
     title={{VBench}: Comprehensive Benchmark Suite for Video Generative Models},
     author={Huang, Ziqi and He, Yinan and Yu, Jiashuo and Zhang, Fan and Si, Chenyang and Jiang, Yuming and Zhang, Yuanhan and Wu, Tianxing and Jin, Qingyang and Chanpaisit, Nattapol and Wang, Yaohui and Chen, Xinyuan and Wang, Limin and Lin, Dahua and Qiao, Yu and Liu, Ziwei},
     booktitle={CVPR},
     year={2024}
 }

@inproceedings{latentreframe,
  title={Latent-reframe: Enabling camera control for video diffusion models without training},
  author={Zhou, Zhenghong and An, Jie and Luo, Jiebo},
  booktitle={ICCV},
  pages={12779--12789},
  year={2025}
}

@article{pi,
  title={Permutation-Equivariant Visual Geometry Learning},
  author={Wang, Yifan and Zhou, Jianjun and Zhu, Haoyi and Chang, Wenzheng and Zhou, Yang and Li, Zizun and Chen, Junyi and Pang, Jiangmiao and Shen, Chunhua and He, Tong},
  journal={arXiv preprint arXiv:2507.13347},
  year={2025}
}

@article{dinov2,
  title={Dinov2: Learning robust visual features without supervision},
  author={Oquab, Maxime and Darcet, Timoth{\'e}e and Moutakanni, Th{\'e}o and Vo, Huy and Szafraniec, Marc and Khalidov, Vasil and Fernandez, Pierre and Haziza, Daniel and Massa, Francisco and El-Nouby, Alaaeldin and others},
  journal={arXiv preprint arXiv:2304.07193},
  year={2023}
}

@inproceedings{clip,
  title={Learning transferable visual models from natural language supervision},
  author={Radford, Alec and Kim, Jong Wook and Hallacy, Chris and Ramesh, Aditya and Goh, Gabriel and Agarwal, Sandhini and Sastry, Girish and Askell, Amanda and Mishkin, Pamela and Clark, Jack and others},
  booktitle={ICML},
  year={2021}
}

@article{fu2025trajmaster,
  title={{3DTrajMaster}: Mastering {3D} Trajectory for Multi-Entity Motion in Video Generation},
  author={Fu, Xiao and Liu, Xian and Wang, Xintao and Peng, Sida and Xia, Menghan and Shi, Xiaoyu and Yuan, Ziyang and Wan, Pengfei and Zhang, Di and Lin, Dahua},
  journal={arXiv preprint arXiv:2412.07759},
  year={2024}
}

@article{carion2025sam3,
  title={{SAM 3}: Segment Anything with Concepts},
  author={Carion, Nicolas and Gustafson, Laura and Hu, Yuan-Ting and Debnath, Shoubhik and Hu, Ronghang and Suris, Didac and Ryali, Chaitanya and Alwala, Kalyan Vasudev and Khedr, Haitham and Huang, Andrew and others},
  journal={arXiv preprint arXiv:2511.16719},
  year={2025}
}

@inproceedings{wang2026frame,
  title={{Frame In-N-Out}: Unbounded Controllable Image-to-Video Generation},
  author={Wang, Boyang and Chen, Xuweiyi and Gadelha, Matheus and Cheng, Zezhou},
  booktitle={Advances in Neural Information Processing Systems},
  volume={38},
  pages={3291--3324},
  year={2025}
}

@article{wang2025customx,
  title={CustomX: Unified Character, Action, and Scene Customization in Video World Models},
  author={Wang, Yitong and Wei, Fangyun and Zhang, Hongyang and Dai, Bo and Lu, Yan},
  journal={arXiv preprint arXiv:2512.17796},
  year={2025}
}

@article{ko20263dreambooth,
  title={3DreamBooth: High-Fidelity 3D Subject-Driven Video Generation Model},
  author={Ko, Hyun-kyu and Park, Jihyeon and Kim, Younghyun and Park, Dongheok and Park, Eunbyung},
  journal={arXiv preprint arXiv:2603.18524},
  year={2026}
}

@article{song2026mv,
  title={Mv-s2v: Multi-view subject-consistent video generation},
  author={Song, Ziyang and Gong, Xinyu and Liu, Bangya and Zhao, Zelin},
  journal={arXiv preprint arXiv:2601.17756},
  year={2026}
}

\appendix
\newpage
In the supplementary material, we provide additional qualitative results (\cref{supp:results}), details of the offline motion-control interface (\cref{supp:gui}), implementation details and inference cost (\cref{supp:details}), additional baseline comparisons (\cref{supp:baselines}), 3D consistency analysis (\cref{supp:3d_all}), and failure-case analysis (\cref{supp:limitation}).

\section{More Visualization Results}
\label{supp:results}

We provide more results of our method to illustrate its performance. These results include cases from games, movies, and real-world cases. Video results are available on the \href{https://zhouzhenghong-gt.github.io/Tri-Prompting-Page/}{project page}.

\begin{figure*}[t]
    \centering
    \includegraphics[width=1.0\linewidth]{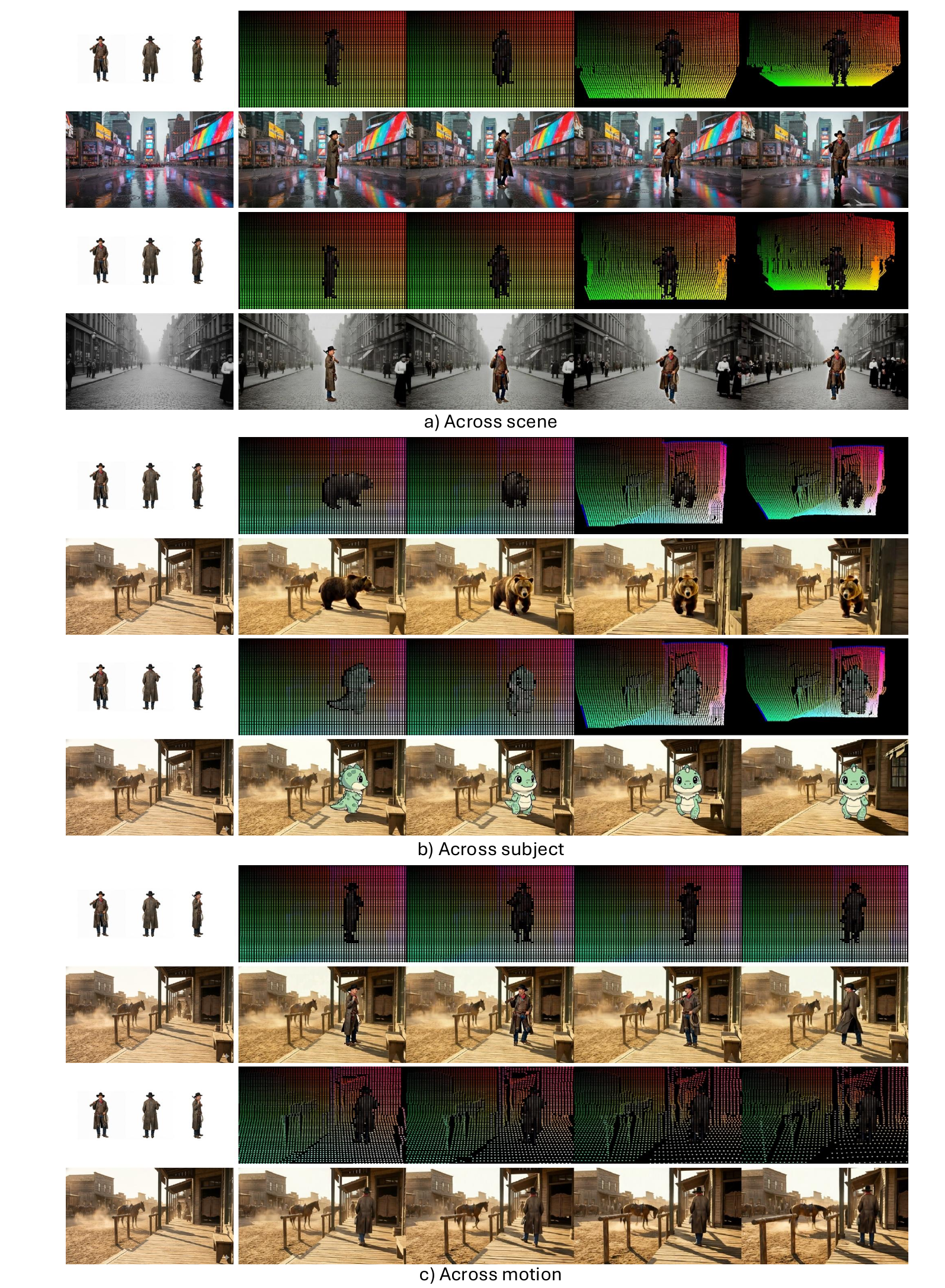}
    \caption{\textbf{Subject insertion applications across scene, subject, and motion.} For each example, the left column shows the multi-view subject reference images (top) and the original background image (bottom). The top row visualizes the motion-control video. In the bottom row, the first frame is the scene conditioning image, followed by frames generated by our method.}
    \label{fig:cowboy}
\end{figure*}

In \cref{fig:cowboy}, we use a Cowboy video as the base to demonstrate representative controls across scene, subject, and motion. When the background scene is changed to a Times Square /old street and the subject is replaced by characters like a bear or anime character, the results successfully maintain the multi-view appearance of the subject and harmonious subject-scene interaction effects, including shadow casting and smooth motion (e.g., movement and turning). The final two rows illustrate object motion control (a $360^\circ$ in-place rotation) and camera control (moving forward). These results demonstrate the robustness of our method, confirming its ability to adhere to all input conditions (scene image, multi-view subject, and motion) and to generate high-quality video with realistic interaction effects across them.

\begin{figure*}[t]
    \centering
    \includegraphics[width=1.0\linewidth]{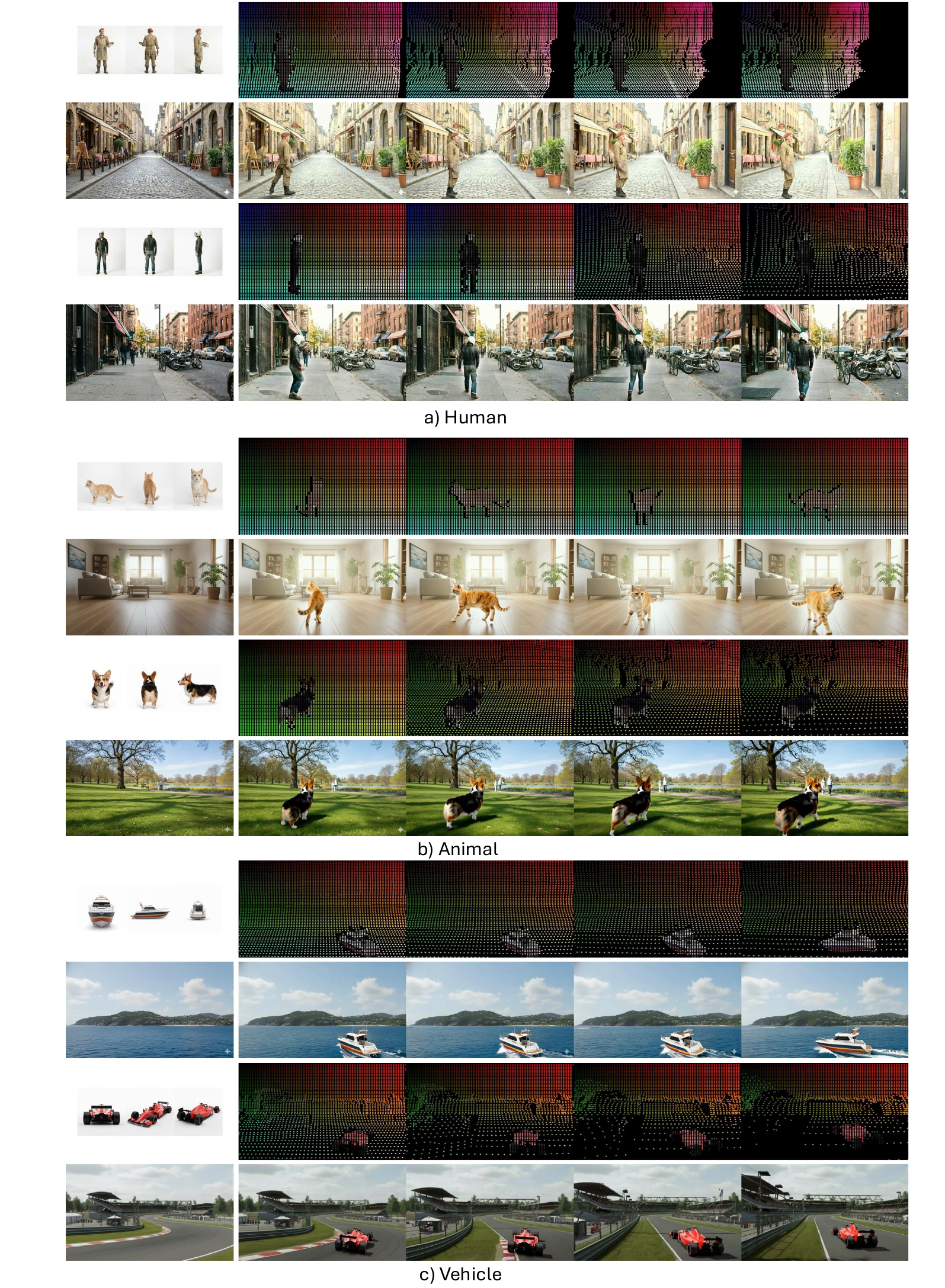}
    \caption{\textbf{Subject insertion applications in real-world cases.} Covering both non-rigid subjects (humans/animals) and rigid objects (vehicles). In both categories, our approach achieves high video quality while following the motion instructions.}
    \label{fig:real}
\end{figure*}

\cref{fig:real} presents subject insertion applications in real-world cases, covering humans, animals, and vehicles. The results show that our method remains effective beyond stylized domains, producing high-quality inserted subjects that are consistent with the input appearance and motion while blending naturally into real scenes.

\begin{figure*}[t]
    \centering
    \includegraphics[width=1.0\linewidth]{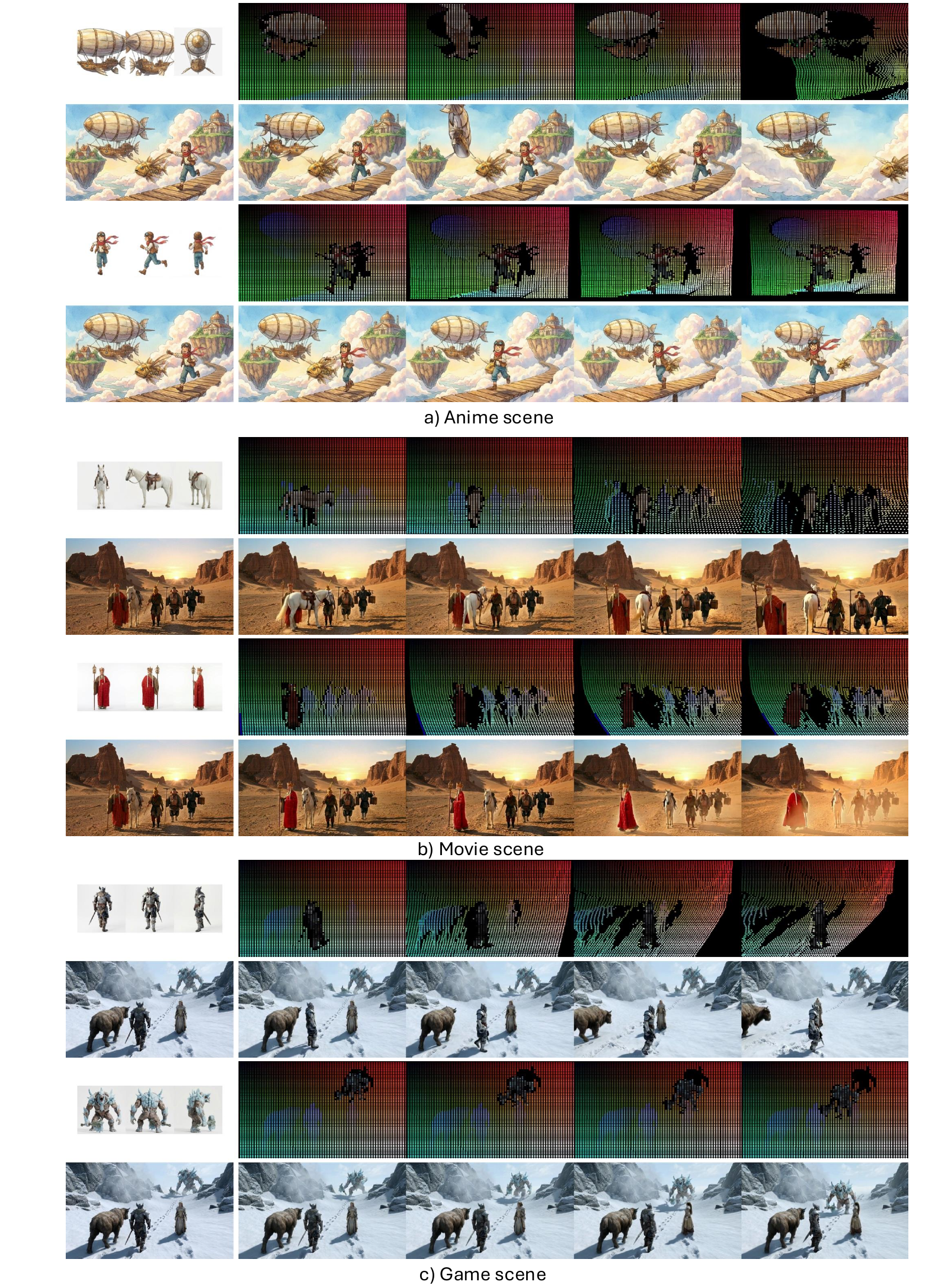}
    \caption{\textbf{Subject manipulation applications in anime, movie, and game scenes.} Generalization results across fictional domains, confirming our model's ability to synthesize diverse and varied video output.}
    \label{fig:sam3d}
\end{figure*}

\cref{fig:sam3d} focuses on subject manipulation applications and shows that the method generalizes well across diverse fictional domains, including anime, movie, and game scenes.
The generated videos align with the given subject images, preserving details (e.g., the horse and man during turnarounds). They also track the input motion while generating realistic interaction effects in various environments.

\section{Offline Motion-Control Video Construction}\label{supp:gui}

\begin{figure}[htbp]
    \centering
    \includegraphics[width=\linewidth]{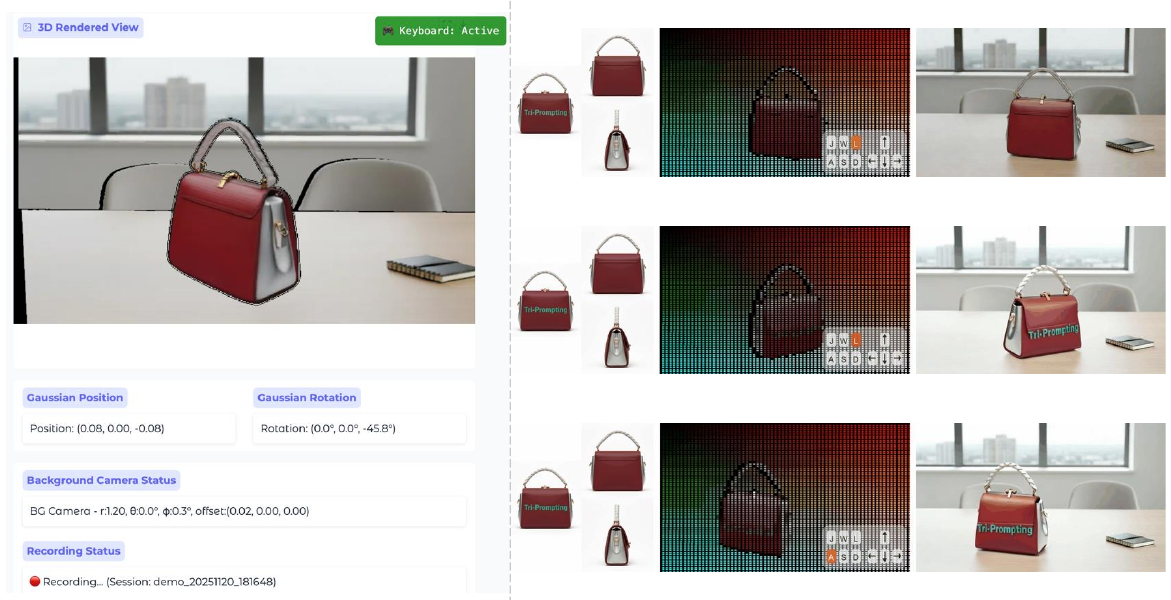}
    \caption{\textbf{Offline control interface.} Left: Interface for specifying camera and subject transformations. Right: Examples of subject images, motion control video with keyboard, and generated video.}
    \label{fig:gui}
    \vspace{-3mm}
\end{figure}

As mentioned in \cref{method:inference_workflow} in the main paper and shown in \cref{fig:gui}, we design an offline control interface to collect user motion instructions. The user selects a background image and a 3D-aware subject, then controls the background/camera and the 3D subject via the keyboard. We composite the background with the 3D subject's rendering and convert the background into a \(70\times70\) XYZ point and the subject rendering into a low-resolution RGB point, as stated in the main paper. The interface records the manipulation session; upon completion, the recording is temporally resampled to 49 frames and used as the model's motion control video.

\section{Implementation Details}\label{supp:details}

\begin{wraptable}[18]{r}{0.48\textwidth}
\vspace{-3mm}
\centering
\caption{\textbf{Detailed hyperparameter settings.}}
\resizebox{\linewidth}{!}{
\setlength\tabcolsep{6pt}
\small
\begin{tabular}{cc}
\toprule
\textbf{Parameter} & \textbf{Value} \\
\midrule
Video frames & $49$ \\
Spatial resolution & $832{\times}480$ \\
Denoising steps & $50$ \\
Base model & Phantom 14B \\
\hdashline
Stage-I Learning rate & $1e-4$ \\
Stage-I batch size & $8$ \\
Stage-I step & $2500$ \\
Stage-I LoRA rank & $64$ \\
\hdashline
Stage-II Learning rate & $1e-4$ \\
Stage-II batch size & $32$ \\
Stage-II step & $2074$ \\
Stage-II ControlNet layers & $18$ \\
\hdashline
ControlNet decay steps $N_{\text{decay}}$ & $10$ \\
Minimum control scale $s_{min}$ & $0.005$ \\
\bottomrule
\end{tabular}
}
\label{tab:hyperparameters}
\vspace{-3mm}
\end{wraptable}
\cref{tab:hyperparameters} summarizes the training and inference hyperparameters of Tri-Prompting, including the training data, model configuration, and the settings used in the two different training stages. In particular, Stage I fine-tunes the base Phantom 14B model with LoRA for scene and multi-view subject control, while Stage II further trains the ControlNet module for dual-conditioning motion control. We leave joint optimization of LoRA and ControlNet to future work. We also report the inference hyperparameters for the ControlNet scale schedule, namely the decay steps $N_{\text{decay}}$ and the minimum control scale $s_{\min}$, which correspond to the inference-time ControlNet scale scheduling strategy described in \cref{method:inference_aug}.
Unless otherwise specified, this schedule is used in both the comparison with Phantom and the application results. We do not apply this strategy in the comparison with DaS, since in that setting both the low-resolution RGB guidance and the XYZ tracking points are directly derived from the original video.

\subsection{Inference Cost}
\label{supp:cost}
We measure Phantom, Stage I, and Stage II on the same eight A100-80GB GPUs, jointly generating one 49-frame, $832\times480$ video with 50 denoising steps. \Cref{tab:inference_cost} reports results on the same three cases. Timing includes input loading, distributed diffusion, VAE decoding, and video encoding; one-time model loading is excluded. Stage I adds 7.10\% wall time over Phantom, and Stage II adds 46.75\% over Stage I while enabling foreground/background motion control.

\begin{table}[htbp]
\centering
\caption{\textbf{Inference cost on eight A100-80GB GPUs.} Wall time is mean $\pm$ population standard deviation over three cases. Peak memory is allocated/reserved GiB per GPU.}
\label{tab:inference_cost}
\small
\setlength{\tabcolsep}{5pt}
\begin{tabular}{@{}lcccc@{}}
\toprule
Method & Wall time (s) & Min/video & Total GPU-min & Peak memory \\
\midrule
Phantom & $147.122\pm0.109$ & 2.452 & 19.616 & 38.578 / 43.252 \\
Ours, Stage I & $157.566\pm0.118$ & 2.626 & 21.009 & 39.174 / 42.928 \\
Ours, Stage II & $231.235\pm0.315$ & 3.854 & 30.831 & 52.616 / 54.740 \\
\bottomrule
\end{tabular}
\end{table}

\section{Additional Baseline Comparisons}
\label{supp:baselines}

\subsection{OmniVCus Reference Inputs}
\label{supp:omnivcus}
We evaluate the released OmniVCus~\cite{cai2025omnivcus} checkpoint with text and foreground-mask videos on the same 106 cases and three seeds. The single-reference setting uses its native image-reference input; the three-view setting supplies a fixed montage through the same input. Scene images and camera trajectories are not supplied separately. \Cref{tab:omnivcus_refs} reports both settings alongside Tri-Prompting, which uses the first-frame scene image, three subject views, and RGB/XYZ motion cues. The montage improves DINO identity over a single reference; Tri-Prompting has higher scores on all six metrics.

For the 3D consistency comparison in \cref{tab:main_results}(b), we use 18 rigid cases with six outputs per case (108 outputs per setting). We apply each output's foreground mask before $\pi^3$ reconstruction and align the point clouds with a shared-frame Sim(3) transform. We average $E_{\text{align}}$ over the 84 outputs successfully reconstructed across eight settings: Phantom, Stage I, Stage II with one, two, or three references (including both scale schedules for three references), and the two OmniVCus inputs. Outputs are matched by case and fixed within-case output rank.

\begin{table}[htbp]
\centering
\caption{\textbf{OmniVCus reference-input comparison.} 318 outputs per setting; higher is better.}
\label{tab:omnivcus_refs}
\small
\setlength{\tabcolsep}{4pt}
\begin{tabular}{@{}lcccccc@{}}
\toprule
Method & Aes. & Smooth & Dynamic & Imaging & $\mathcal{I}_{\text{dino}}$ & $\mathcal{I}_{\text{clip}}$ \\
\midrule
OmniVCus (1-ref) & 0.619 & 0.987 & 0.619 & 0.707 & 0.695 & 0.845 \\
OmniVCus (3-view montage) & 0.618 & 0.987 & 0.613 & 0.703 & 0.703 & 0.846 \\
Ours (3-view) & \textbf{0.665} & \textbf{0.989} & \textbf{0.903} & \textbf{0.715} & \textbf{0.746} & \textbf{0.862} \\
\bottomrule
\end{tabular}
\end{table}

\subsection{Motion and Identity Evaluation}
\label{supp:360motion}
We use the single-entity camera-C12 subset of 360Motion~\cite{fu2025trajmaster}, held out from 3DTrajMaster's training on the other 11 cameras. Tri-Prompting is evaluated without training on this dataset. We use the released root 6DoF controls for 3DTrajMaster and downsample the released videos to construct Tri-Prompting's low-resolution foreground RGB. On our 106-case set, 3DTrajMaster receives matched global/entity text and 13 root-6DoF tokens derived from recorded position and rotation. Phantom receives the scene image, three subject references, and text, without motion inputs.
Our 106 cases comprise 81 dynamic and 25 static backgrounds; subject controls include 59 rotation-and-translation, 38 rotation-only, and 9 translation-only cases.

We extract target and generated subject masks using SAM~3~\cite{carion2025sam3} on public 360Motion and SAM~2~\cite{sam2} on our evaluation set. Bbox IoU is the intersection area divided by the union area of the tight bounding boxes enclosing the target and generated subject masks. Mask IoU applies the same ratio to the masks themselves. Both metrics are averaged over valid frames. Multi-view identity uses the DINO/CLIP similarity defined in the main text. We evaluate two fixed ControlNet settings across both datasets: constant scale $1.0$, or decay from $1.0$ to $0.005$ over the first 10 of 50 denoising steps.

\subsection{Comparison with MotionBooth}
\label{supp:motionbooth}
MotionBooth~\cite{wu2024motionbooth} uses three subject references and masks, 300-step per-subject personalization, and target 2D boxes. We evaluate its public Zeroscope implementation on the same 106 cases and three seeds; Tri-Prompting and Phantom use the inputs described in \cref{supp:360motion}. To match MotionBooth's native setting, all videos are normalized to 24 frames at $576\times320$ and 10 fps. \Cref{tab:motionbooth} reports quality and identity under this protocol. With the same reference count, Tri-Prompting improves $\mathcal{I}_{\text{dino}}$ by 0.175 and $\mathcal{I}_{\text{clip}}$ by 0.091 over MotionBooth.

\begin{table}[htbp]
\centering
\caption{\textbf{Comparison with MotionBooth.} 318 outputs per method, evaluated at 24 frames, $576\times320$, and 10 fps; higher is better.}
\label{tab:motionbooth}
\small
\setlength{\tabcolsep}{4pt}
\begin{tabular}{@{}lcccccc@{}}
\toprule
Method & Aes. & Smooth & Dynamic & Imaging & $\mathcal{I}_{\text{dino}}$ & $\mathcal{I}_{\text{clip}}$ \\
\midrule
MotionBooth-public & 0.540 & \textbf{0.984} & 0.349 & 0.637 & 0.562 & 0.753 \\
Phantom & 0.625 & 0.982 & 0.613 & 0.670 & 0.715 & 0.839 \\
Ours & \textbf{0.647} & 0.983 & \textbf{0.918} & \textbf{0.673} & \textbf{0.737} & \textbf{0.844} \\
\bottomrule
\end{tabular}
\end{table}

\clearpage
\section{3D Consistency Across All Subjects}
\label{supp:3d_all}
The rigid-subject evaluation in \cref{tab:main_results}(b) uses shared Sim(3) alignment to static references. For non-rigid subjects, legitimate articulation such as walking also contributes to the alignment error. We extend $\pi^3$ reconstruction to all 106 cases and three seeds, including dynamic subjects. \Cref{tab:3d_all} reports reconstruction success over all 318 outputs and $E_{\text{align}}$ over the 230 outputs successfully reconstructed by all three methods.

\begin{table}[htbp]
\centering
\caption{\textbf{3D reconstruction across all subjects.} Alignment error is averaged over the 230 common successes.}
\label{tab:3d_all}
\small
\setlength{\tabcolsep}{8pt}
\begin{tabular}{@{}lcc@{}}
\toprule
Method & Reconstruction successes & $E_{\text{align}}\downarrow$ \\
\midrule
Phantom & 264/318 & 0.062 \\
Ours, Stage I & 282/318 & 0.079 \\
Ours, Stage II & \textbf{298/318} & \textbf{0.057} \\
\bottomrule
\end{tabular}
\end{table}

Stage II has the highest reconstruction success and lowest alignment error on the shared subset. For dynamic subjects, the multi-view identity metrics in \cref{tab:main_results}(b) complement geometry by assessing semantic and appearance consistency.

\section{Limitation}
\label{supp:limitation}

\begin{figure}[htbp]
    \centering
    \includegraphics[width=0.7\linewidth]{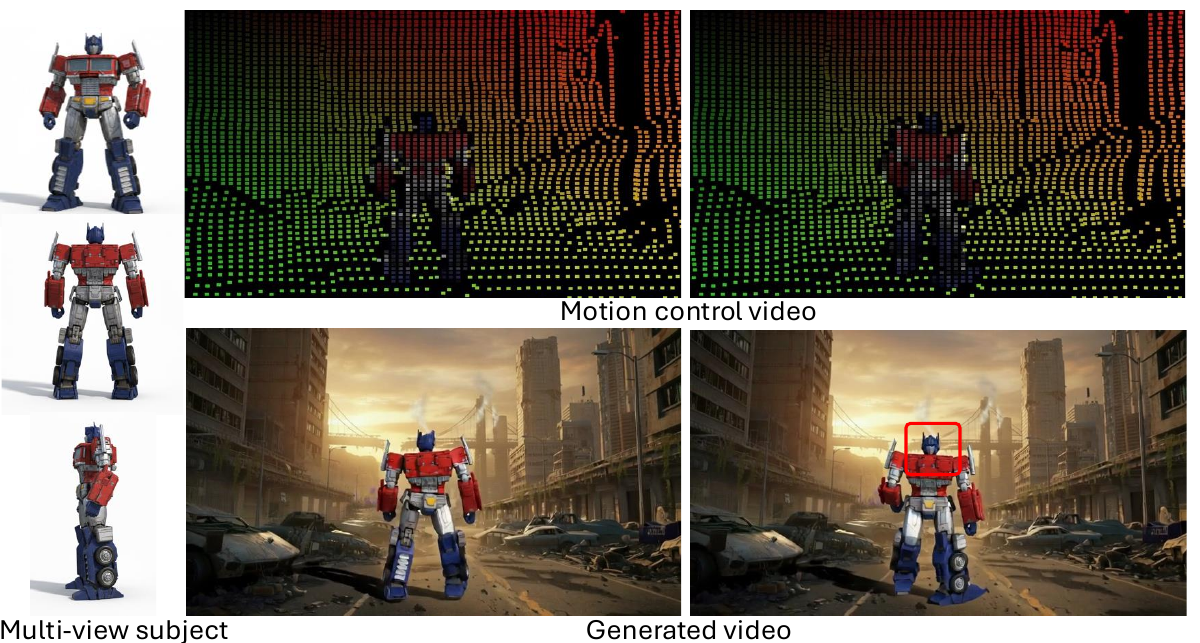}
    \caption{\textbf{A failure case of Our Method.}}
    \label{fig:limitation}
\end{figure}

While Tri-Prompting is generally robust under diverse controls, \cref{fig:limitation} illustrates a failure case caused by ambiguity in highly symmetric multi-view references. In this example, the subject's front and back views share very similar appearance, making the identity orientation under motion less distinguishable. As a result, the generated video may exhibit a short-lived inconsistency, such as a transient flip in head orientation during turning. This issue is not persistent across the whole sequence, but it reveals that the current model can struggle when the reference set provides insufficient asymmetric cues. In practice, this failure mode can often be alleviated by using more distinctive reference views or adding stronger motion constraints. In addition, the current method focuses on one primary controllable subject and does not support complex multi-subject generation.

\end{document}